\documentclass[review]{elsarticle}

\usepackage{lineno,hyperref}

\usepackage{algorithm}
\usepackage{algorithmicx}
\usepackage{algpseudocode}
\usepackage{amsmath}
\usepackage{booktabs}  
\usepackage{multirow}
\usepackage{graphicx}
\usepackage{bm}
\usepackage{xcolor}
\usepackage{makecell}
\usepackage{expl3}
\usepackage{tabularx}
\usepackage{enumitem}
\usepackage{comment}
\usepackage{amsfonts}
\usepackage{subfigure}
\newcommand{\etal}{\textit{et al}.~}
\newcommand{\ie}{\textit{i}.\textit{e}.~}

\newcommand{\eg}{\textit{e}.\textit{g}.~}

\modulolinenumbers[5]

\journal{Journal of \LaTeX\ Templates}

\begin{document}

\begin{frontmatter}

\title{CASINet: Content-Adaptive Scale Interaction Networks for Scene Parsing}

\fntext[myfootnote]{This work was done when Xin and Zhizheng were interns at Microsoft Research Asia.}

\author[mymainaddress]{Xin Jin}

\author[mysecondaryaddress]{Cuiling Lan\corref{mycorrespondingauthor}}
\cortext[mycorrespondingauthor]{Corresponding author}
\ead{culan@microsoft.com}

\author[mysecondaryaddress]{Wenjun Zeng}

\author[mymainaddress]{Zhizheng Zhang}

\author[mymainaddress]{Zhibo Chen\corref{mycorrespondingauthor}}

\address[mymainaddress]{University of Science and Technology of China, Hefei, Anhui, 230026, China}
\address[mysecondaryaddress]{Microsoft Research Asia, No. 5 Dan Ling Street, Haidian District, Beijing, 100080, China}

\begin{abstract}
Objects at different spatial positions in an image exhibit different scales. Adaptive receptive fields are expected to capture suitable ranges of context for accurate pixel level semantic prediction. Recently, atrous convolution with different dilation rates has been used to generate features of multi-scales through several branches which are then fused for prediction. However, there is a lack of explicit interaction among the branches of different scales to adaptively make full use of the contexts. In this paper, we propose a Content-Adaptive Scale Interaction Network (CASINet) to exploit the multi-scale features for scene parsing. We build CASINet based on the classic Atrous Spatial Pyramid Pooling (ASPP) module, followed by a proposed contextual scale interaction (CSI) module, and a scale adaptation (SA) module. Specifically, in the CSI module, for each spatial position of some scale, instead of being limited by a fixed set of convolutional filters that are shared across different spatial positions for feature learning, we promote the adaptivity of the convolutional filters to spatial positions. We achieve this by the context interaction among the features of different scales. The SA module explicitly and softly selects the suitable scale for each spatial position and each channel. Ablation studies demonstrate the effectiveness of the proposed modules. We achieve state-of-the-art performance on three scene parsing benchmarks Cityscapes, ADE20K and LIP.
\end{abstract}


\end{frontmatter}


\section{Introduction}

Scene parsing or semantic segmentation is a fundamental and challenging task. The purpose is to predict the semantics of each pixel including stuffs (e.g. sky, road) and objects (e.g. person, car). This task has recently attracted remarkable attention for both images \cite{lateef2019survey,wang2019aleatoric,li2020weakly,zhang2020mfenet,sun2019high,li2019global} and video \cite{tian2016video,wang2016transductive,sun2016interactive,zhu2019improving,zhang2019refined}, and has benefited many important applications in computer vision, such as autonomous driving \cite{wang2020deep}, robot sensing \cite{zhang2018road,park2018meet}, and image editing/captioning \cite{xie2019weakly,zhao2019multimodal}.



\begin{figure}
  \centering
  \includegraphics[width=0.7\linewidth]{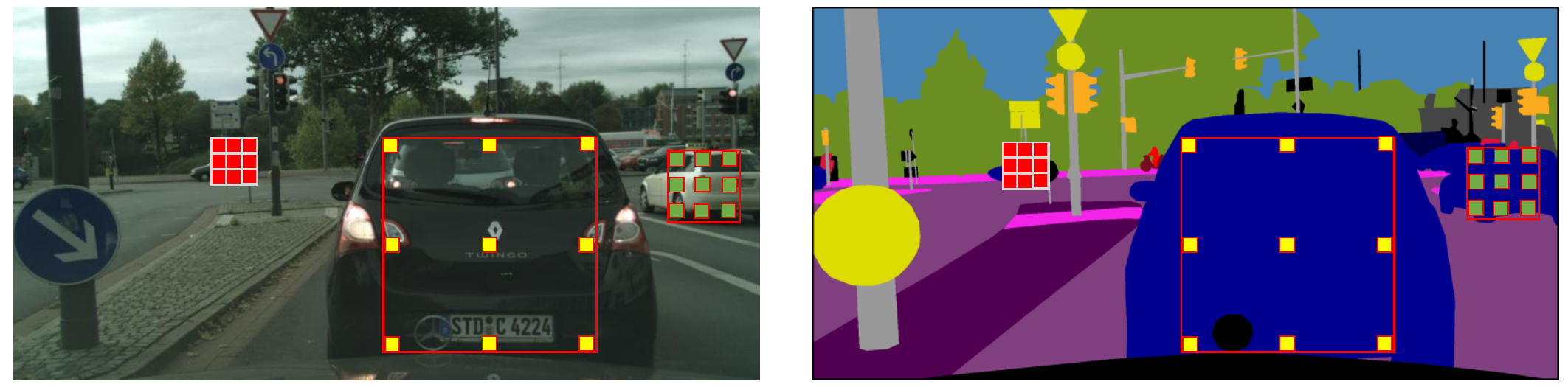}
  \caption{Example of the scale variations of objects in a street scene. The same category of objects, such as cars, may vary largely in scale. Each pixel needs a suitable size receptive field to catch the best context.}
  \label{fig:ill}
  \vspace{-3mm}
\end{figure}

With the development of the Fully Convolutional Network (FCN) \cite{long2015fully}, semantic image segmentation has achieved promising results with significantly improved feature representation. However, as shown in Figure \ref{fig:ill}, objects within a picture are typically diverse on scales. The CNNs with standard convolutions can not handle diverse scales due to fixed receptive fields. Objects that are larger than the receptive fields often have inconsistent parsing prediction while objects that are smaller than the receptive fields are often ignored/mislabeled~\cite{zhang2017scale}.


To deal with scale diversity, multi-scale context fusion schemes have been proposed \cite{zhao2017pyramid,chen2018deeplab,yang2018denseaspp,li2019global}. The early representative work is the pyramid pooling module (PPM) in PSPNet \cite{zhao2017pyramid}, which pools the feature map at multiple rates and multiple effective field-of-views, capturing objects at multiple scales\cite{chen2017rethinking}. Atrous spatial pyramid pooling (ASPP) \cite{chen2018deeplab,yang2018denseaspp} modules use atrous/dilated convolutions \cite{chen2014semantic} to effectively enlarge the field of view of filters to have multi-scale features. Atrous/Dilated convolutions with different atrous rates lead to different receptive field sizes in generating features. All the above approaches concatenate these features followed by convolution operations to exploit multi-scale contexts. To some extent, these convolution filters promote the interactions and selections among features of different scales. However, after training, the filter weights are fixed and thus the interaction and selection are not highly content adaptive and flexible. For different spatial positions, the object/stuff categories and scales are different. The feature interaction among scales for exploiting contextual features of different scales should be more content adaptive rather than position invariant. Moreover, for each scale, the convolutional filters are shared across different spatial positions. Ideally, the weights of the filters should adapt to the varied contents on different spatial positions. 

\begin{figure*}
  \includegraphics[width=0.99\textwidth]{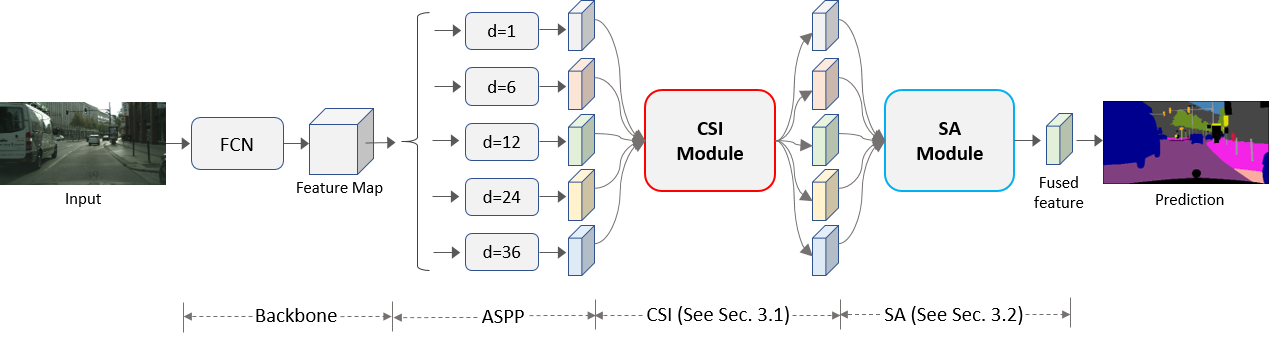}
  \caption{An overview of the Content-Adaptive Scale Interaction Network (CASINet). We build it based on the classic Atrous Spatial Pyramid Pooling (ASPP) \cite{chen2017rethinking}, followed by the proposed Contextual Scale Interaction (CSI) module, and the Scale Adaptation (SA) module. The CSI module mimics the function of spatial adaptive filtering for better feature learning/refinement. The SA module adaptively combines/fuses the features of different scales for objects/stuffs of different sizes. The fused feature is fed to a convolution layer followed by SoftMax for prediction. Note $d$ denotes the dilation rate of the atrous convolution.}
  \label{fig:pipeline}
\end{figure*}

To address the above problems, we propose a Content-Adaptive Scale Interaction Network (CASINet) to adaptively exploit multi-scale features for scene parsing. Figure~\ref{fig:pipeline} shows the overall flowchart of our framework. We build our framework on top of the classic Atrous Spatial Pyramid Pooling (ASPP) module, which provides $K$ ($K=5$ in our design) feature maps, with each obtained from the filtering by a set of fixed size convolutional filters. 

\begin{figure}
  \centering
  \includegraphics[width=0.7\linewidth]{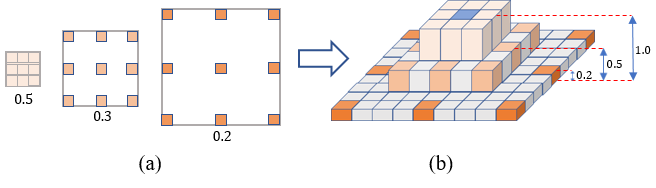}
  \vspace{-5mm}
  \caption{Illustration of the generation of a new filter from three filters by combing them with weights of, e.g., 0.5, 0.3, 0.2.}
  \vspace{-2mm}
  \label{fig:RF}
\end{figure}

\textbf{First}, we design a contextual scale interaction (CSI) module for feature refinement, which intends to have adaptive filters for features at each scale. As illustrated in Figure \ref{fig:RF}, by weighted combining several (\eg, three) filters of different sizes/dilation rates, we can have a new filter, where the combination weights control the shape of the filter. Motivated by this, since the features of different scales are obtained by using corresponding filters of different sizes, we can re-generate a refined feature by weighted combining the multi-scale features to mimic the use of adaptive filters for each spatial position of each scale. \emph{Intuitively, the combination could consider the correlations among the responses of different filters to downplay the features from weakly correlated scales.} We achieve this goal by leveraging the non-local operations \cite{wang2018non} across the $K$ scales of features. For the feature of a scale at a spatial position, we compute the refined response as a weighted sum of the features at all the scales of the same spatial position. For different spatial positions, the combination weights are different to be content adaptive. Note that even though the filter kernel weights also aim at capturing the varied correlations within a receptive field, since the filter kernel weights are shared across the different spatial positions of an image, they cannot meet the varied requirements of different spatial positions. \textbf{Second}, we design a spatial and channel adaptive scale adaptation (SA) module to adaptively combine the appropriate features of different scales for objects/stuffs of different sizes. For each spatial position, we simultaneously learn scale and channel attention. 

Our main contributions are summarized as follows:
\begin{itemize}[leftmargin=*,noitemsep,nolistsep]
    \item We propose a simple yet effective content-adaptive scale interaction network (CASINet) to efficiently exploit the multi-scale features for scene parsing. This can mimic spatial adaptive filtering for each branch/scale for feature refinement. 
    
    \item We propose a contextual scale interaction (CSI) module which enables the function of spatial adaptive filtering for better feature learning, by leveraging the scale context interaction. In addition, we propose a scale adaptation (SA) module that facilitates the appropriate scale selection and combination adapted to spatial positions and channels.  
    
\end{itemize}

We validate the effectiveness of the proposed modules through the ablation studies. We achieve state-of-the-art results on three scene parsing benchmarks including Cityscapes \cite{cordts2016cityscapes}, ADE20K \cite{zhou2017scene}, and LIP datasets \cite{gong2017look}.

\section{Related Work}

\textbf{Multi-Scale Feature Exploration.} Scene parsing has achieved great progress with the development of Fully Convolutional Networks (FCNs) \cite{long2015fully}. To alleviate the local receptive field (RF) issue of convolution operations in FCNs, several network variants are proposed to generate and aggregate multi-scale features. Motivated by the spatial pyramid matching \cite{lazebnik2006beyond}, PSPNet \cite{zhao2017pyramid}, Deeplabv2 \cite{chen2018deeplab}, and Deeplabv3 \cite{chen2017rethinking} are proposed to concatenate features of multiple receptive field sizes together for the semantic prediction. PSPNet \cite{zhao2017pyramid} employs four spatial pyramid pooling (down-sampling) layers in parallel, named pyramid pooling module (PPM), to aggregate information from multiple receptive field sizes. Deeplabv2 \cite{chen2018deeplab} and Deeplabv3 \cite{chen2017rethinking} both adopt atrous spatial pyramid pooling (ASPP) to concatenate features from multiple atrous convolution layers with different dilation rates arranged in parallel. Based on such ASPP structure, some closely related work in other fields, such as salient edges \cite{song2018pyramid} and salient object detection \cite{wang2019salient} are also explored to leverage multi-scale features for fine-grained feature representation learning. Considering the simple structure yet superior performance of ASPP \cite{chen2017rethinking,chen2018deeplab}, our design follows the multi-branch structure as used in ASPP.

Besides the above popular multi-branch structures, there are some other works that exploit spatial contexts to achieve the effect of different receptive field sizes. Peng \etal~  \cite{peng2017large} enlarge the kernel size with a decomposed structure for global convolution. Zhang \etal \cite{zhang2017scale} propose a scale-adaptive convolution to acquire flexible-size receptive fields by adding a scale regression layer. Lin, \etal \cite{lin2017refinenet,ding2018context} utilize the encoder-decoder structure to fuse middle-level and high-level features.
DAG-RNN \cite{shuai2018scene} employs a recurrent neural network to capture the contextual dependencies over local features. EncNet \cite{zhang2018context} introduces a global encoding layer to learn whether some category of object/stuff exists and uses this as channel attention over the score maps.

However, for the set of works~\cite{zhao2017pyramid,chen2017rethinking,chen2018deeplab,sun2019high,liu2019auto}, there is a lack of explicit \emph{content adaptive interaction} among the features of different scales. Moreover, for each scale, the set of filters is fixed and invariant to different spatial positions. Intuitively, since the correlations vary for different distances and contents, flexible filters are desired. Even though the filter kernel weights also aim at reflecting such varying correlations, the filter kernel weights are shared across  different spatial positions and thus cannot meet the varied requirements of the different spatial positions. In contrast, our contextual scale interaction (CSI) module achieves this by adaptively adjusting the combination weights of the responses of the filters of different scales, by exploiting the correlations among the responses.


\noindent\textbf{Non-Local Mean.} Non-local mean has proven effective for many tasks, such as image denoising \cite{buades2005non,BM3DTIP07}, and texture synthesis \cite{efros1999texture}. It calculates the pixel/feature value at one position as a weighted sum of all positions to exploit the spatial context. For the convolutional neural networks, Wang \textit{et al.} design a module of non-local block which enables the feature refinement by exploiting the long-range spatial or spatio-temporal contexts \cite{wang2018non}. Fu \etal \cite{fu2019dual} and Yuan \etal \cite{yuan2018ocnet} adopt the non-local block for the spatial information interaction for scene parsing. Fu \etal \cite{fu2019dual} extend the non-local operation to channels which enables the channel information interaction \cite{fu2019dual}. 

In this paper, based on the ASPP \cite{chen2017rethinking,chen2018deeplab} multi-branch framework, we enable the function of spatial adaptive filtering for better feature learning by adaptively combining the responses of filters of different sizes/dilation rates. Since the combination is expected to leverage the  correlations of the contents, we propose to achieve this through the non-local operations among features of different scales. Moreover, we leverage the scale prior, \ie, scale/branch index, as part of the features for better determining the combination weights for the target scale. 

\noindent\textbf{Attention.}
Attention that aims to enhance the important features and suppress the irrelevant features is widely used in many tasks \cite{chorowski2015attention,wu2020dynamic,xu2015show,xu2016ask,wang2017residual,hu2018squeeze,yan2019traffic}. Hu \textit{et al.} \cite{hu2018squeeze} introduce a channel-wise attention mechanism through a squeeze-and-excitation (SE) block to modulate the channel-wise feature responses for image classification. For scene parsing, an attention mechanism is proposed that learns to weight the multi-scale features at each pixel location on the score maps \cite{chen2016attention,kong2018recurrent,kong2019pixel}. For a spatial position at a scale, each attention weight is shared across all channels. Weights are normalized by SoftMax/Gumbel Softmax across scales \cite{chen2016attention,kong2019pixel}. Pang et.al \cite{pang2019towards} propose a feature attention strategy which extracts complementary information (i.e., boundary information) from low-level features to enhance high-level features for precise segmentation. Their purpose is to remedy the complementary boundary-like information to the high-level features. In contrast, ours is an attention across scales at the same high-level features to enable the adaptive scale selection on different spatial positions


Recent attention works mostly focus on developing different neural architectures for obtaining fine-grained feature representations, including using the pyramid structure \cite{li2018pyramid}, asymmetric structure \cite{zhu2019asymmetric}, criss-cross structure \cite{huang2019ccnet}, and with the help of contextual information \cite{zhou2019context}, triplet-supervision (deriving from multi-tasks) \cite{cao2019triply}. In contrast to these methods, we aim to leverage the attention mechanism to develop an adaptive aggregation strategy for cross-scaled feature fusion and selection.


For the nearly same purpose of feature fusion and selection, our scale adaptation (SA) can be viewed as an attention across scales, but it is different from previous works in two aspects. First, we relax the sum 1 constraint (across the scales) over the attention weights. Such constraint is not well aligned with the content based adaptation motivation, as it assumes spatial uniformity and thus makes it hard to adjust the energy for different spatial positions. In contrast, the sum of our attentions of the five scales is allowed to vary over spatial positions, and is more spatially adaptive. Second, we extend the attention to be both channel and spatially adaptive to enable the network to flexibly select/combine the appropriate  scales.   


\section{Proposed CASINet}

We propose a Content-Adaptive Scale Interaction Network (CASINet) to efficiently exploit the multi-scale features for scene parsing. In considering the simple structure yet superior performance of ASPP \cite{chen2017rethinking,chen2018deeplab}, our design follows the multi-branch structure as ASPP. Figure~\ref{fig:pipeline} illustrates the overall framework. For a given input image, the ASPP module generates multi-scale features (\eg~ $K = 5$ scales) from the feature map extracted by a backbone FCN. For the multi-scale features, the proposed contextual scale interaction (CSI) module (see subsection \ref{CSI}) enables spatial adaptive filtering for better feature learning/refinement. Over the refined multi-scale features, the proposed scale adaptation (SA) module (see subsection \ref{SA}) adaptively determines suitable scales for channels and spatial positions to fuse the multi-scale features. The fused feature is fed to a convolution layer followed by SoftMax for prediction. Note that the entire network is end-to-end trained.

\subsection{Contextual Scale Interaction (CSI) Module}\label{CSI}

Objects within an image typically have different sizes. Multiple branch structures like that in ASPP \cite{chen2018deeplab,chen2017rethinking} are usually used to learn features using filters of different sizes/dilation rates in order to adapt to the scales. 
Intuitively, for each scale, since the spatial contents vary, a set of fixed filters is not flexible enough to adapt to the variations over different spatial positions. Thus, spatial adaptive filters/kernels \cite{zhang2017scale,dai2017deformable,jia2016dynamic,su2019pixel} are desired. In addition to that, we go a step further and explore the intersection between different filters/kernels.



We intend to design a module which can mimic the function of spatial adaptive filtering to adapt to varied contents. 
Mathematically, we can represent the response ${\mathbf{y}^{(i,j)}}$ of a new filter at spatial position ($i$,$j$) by weighted combining $K$ filters as
\begin{equation}
  {\mathbf{y}^{(i,j)}}  = \sigma \big( \big(\!\sum_{m=1}^{K} c_{m}^{(i,j)} W_m \big)~  {\mathbf{z}^{(i,j)}} \big),
\end{equation}
where $c_{m}^{(i,j)} $ denotes the combination weights of the $K$ filters, with the $m^{th}$ filter matrix represented by $W_m$, to have a new position adaptive filter $\sum_{m=1}^{K} c_{m}^{(i,j)} W_m$ for spatial position ($i$, $j$).  $\mathbf{z}^{(i,j)}$ denotes the input feature around the position ($i$, $j$). $\sigma(\cdot)$ denotes the activation function. We approximate this by weighted combing the $K$ scale features $\mathbf{x}_{m}^{(i,j)}, m =1,\cdots, K$, which are obtained from filtering operations with different filter sizes, as
\begin{equation}
\begin{split}
  {\mathbf{y}^{(i,j)}} & = \sigma \Big( \big(\!\sum_{m=1}^{K} c_{m}^{(i,j)} W_m \big) ~ {\mathbf{z}^{(i,j)}} \Big) \\
  & \approx \sum_{m=1}^{K} \tilde{c}_{m}^{(i,j)} \sigma \big(W_m   {\mathbf{z}^{(i,j)}}\big) = \sum_{m=1}^{K} \tilde{c}_{m}^{(i,j)} \mathbf{x}_{m}^{(i,j)},
\end{split}
\end{equation}

\begin{figure}[t]
  \centering
  \includegraphics[width=0.7\linewidth]{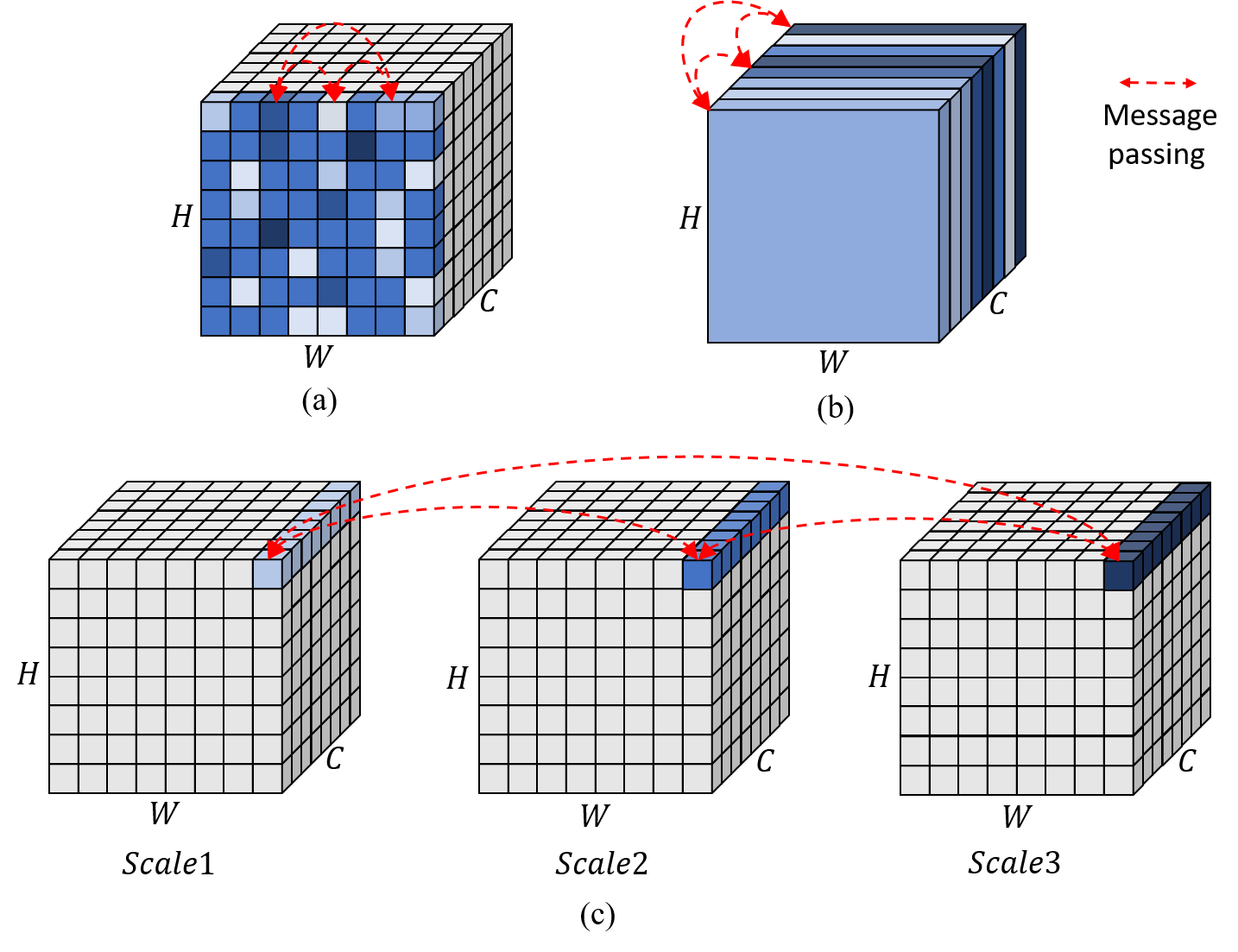}
  \vspace{-5mm}
  \caption{Illustration of different non-local methods. Each cube shows a feature map tensor, with $C$ denoting the number of channels, $H$ and $W$ as the height and width. (a) Spatial-wise non-local operation \cite{yuan2018ocnet,fu2019dual}. (b) Channel-wise non-local operation \cite{fu2019dual}. (c) Proposed scale-wise non-local operation in our CSI module for contextual scale interaction. $Scale1\sim3$ denote different feature tensors corresponding to different scales/receptive fields. }
  \label{fig:dimension}
\end{figure}

In our work, we achieve this by simply enabling the interaction among the multi-scale features and weighted combining them. Like non-local operation \cite{wang2018non}, we model the combination weight by the relation/affinity of the features in an embeded space. Different from previous works which apply non-local operations on spatial (see Figure \ref{fig:dimension}(a)) or channel (see Figure \ref{fig:dimension}(b)), we take it as a tool to achieve scale-wise interaction which enables the spatial adaptive feature combinations across scales, for features of each scale.  

\begin{figure}[t]
  \centering
  \includegraphics[width=0.7\linewidth]{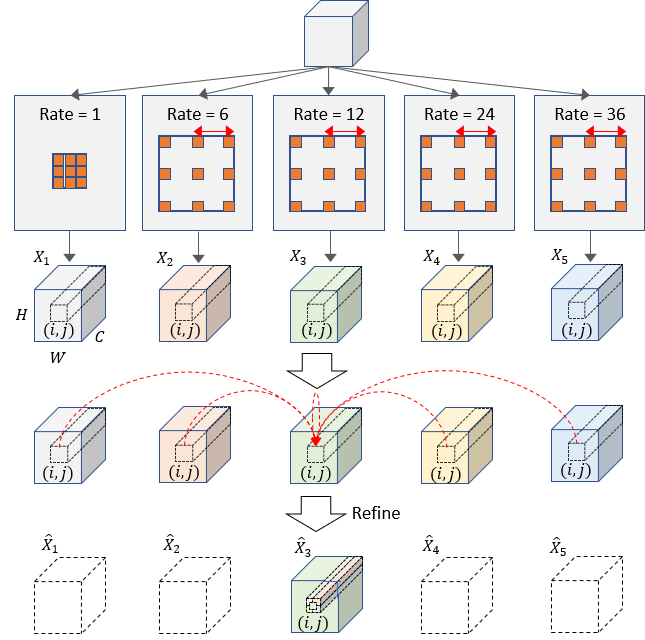}
  \caption{Illustration of the Contextual Scale Interaction (CSI) module through scale-aware non-local operations across scales. For example, we refine the feature of scale 3 at spatial position ($i,j$) by weighted averaging the  features of all the five scales (context) at position ($i,j$) (best viewed in color).}
  \label{fig:SAR}
  \vspace{-2mm}
\end{figure}

Figure~\ref{fig:SAR} illustrates the procedure. Given five ($K=5$) intermediate feature maps (tensors) ${\{X_1, X_2, X_3, X_4, X_5\}}$ from five branches of ASPP with each $X_k \in \mathbb{R}^{H\times W \times C}$ of width $W$, height $H$, and $C$ channels. For each spatial position $(i,j)$, the information interaction is performed across the five scales with each scale being a feature node. The five feature nodes are represented as $\mathbf{x}_{k}^{(i,j)} \in \mathbb{R}^C$, where $k = 1,2,3,4,5$. Then non-local operations are performed for the five features. For the feature of scale $k$, it is calculated as a weighted average of all the scales as
\begin{equation}
  \tilde{\mathbf{x}}_{k}^{(i,j)}  = \frac{1}{\mathcal{C}({\mathbf{x}}^{(i,j)})}\sum_{m=1}^{K} r_{k,m}^{(i,j)}{\mathbf{x}_{m}^{(i,j)}},
\label{eq:refine}
\end{equation}
where $\tilde{c}_{k,m}^{(i,j)} = r_{k,m}^{(i,j)}/{\mathcal{C}({\mathbf{x}}^{(i,j)})}$ denotes the connecting weight in terms of the pairwise affinity, between features $\mathbf{x}_m^{(i,j)}$ (of the source scale $m$) and $\mathbf{x}_k^{(i,j)}$ (of the target scale $k$), $\mathcal{C}({\mathbf{x}}^{(i,j)}) = \sum_{m=1}^{K} r_{k,m}$. Here $r_{k,m}^{(i,j)}$ is obtained by
\begin{equation}
\begin{split}
  r_{k,m}^{(i,j)} = f(\mathbf{x}_k^{(i,j)},k, \mathbf{x}_m^{(i,j)},m)\!=\! e^{\theta_k(\mathbf{x}_{k}^{(i,j)},k)^{T} \phi_m(\mathbf{x}_{m}^{(i,j)},m)}, 
\end{split}
\end{equation}
where $\theta_k(\mathbf{x}_{k}^{(i,j)},k)$, $\phi_m(\mathbf{x}_{m}^{(i,j)},m)$ represent two individual embedding functions implemented by two 1 $\times$ 1 convolutional layers followed by batch normalization (BN) and ReLU activation, these two 1 $\times$ 1 convolutional layers reduce the channel dimensions to the half of origin (which follows the default setting of non-local module \cite{wang2018non}). For each spatial position $(i,j)$, from $r_{k,m}^{(i,j)}$ with $k, m \in \{1,\cdots,K\}$, we can obtain an affinity matrix $G^{(i,j)}$ of $K\times K$ for guiding the interaction. Here, we have 5 different scales, so $K = 5$, as shown in Figure~\ref{fig:SAR}. It is worth noting that we embed the prior information of \textbf{scale index $k$} to better learn the affinity among features of various scales.

Then, we obtain five refined feature maps (tensors), ${\{\hat{X_1}, \cdots, \hat{X_5}\}}$ $\in R^{H\times W \times C}$.

\textbf{Discussion}: Along with the use of non-local blocks \cite{wang2018non} to exploit the long-range spatial or spatio-temporal contexts for feature refinement, the variants of non-local idea in scene parsing \cite{fu2019dual,yuan2018ocnet,han2019new} tend to refine the feature at one spatial/channel position as a weighted sum of all  (spatial/channel) positions within a feature tensor. In contrast, our CSI module exploits the features at the same spatial positions but across different scales to achieve the adaptive information interaction across scales. Figure \ref{fig:dimension} illustrates the (a) spatial-wise, (b) channel-wise, and (c) our scale-wise non-local operations, respectively. To our best knowledge, CSI is the first to exploit the scale contextual information to enable the position adaptive interactions among scales. We will validate that our CSI module is complementary to the function of previous non-local operations in the experiment section.

\subsection{Scale Adaptation (SA) Module}\label{SA}

\begin{figure}[t]
  \centering
  \includegraphics[width=0.7\linewidth]{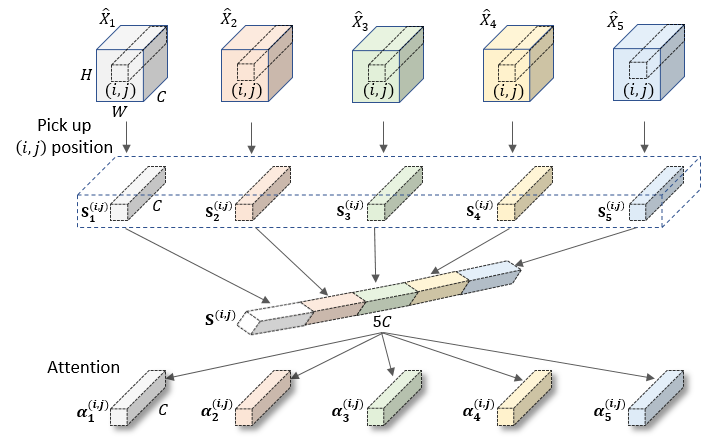}
  \caption{Illustration of our Scale Adaptation (SA) module. To facilitate the illustration, we pick the position ${(i,j)}$ as an example. For each spatial position ${(i,j)}$, we concatenate the features of all scales and learn content adaptive scale attention over each channel of each scale, \ie ~5$\times$C attention values (best viewed in color).}
  \label{fig:SAF}
\end{figure}

We propose a spatial and channel adaptive scale adaptation (SA) module. This enables the adaptive selection of the appropriate receptive fields/scales for different size objects/stuffs. As illustrated in Figure~\ref{fig:SAF}, the refined feature maps (of ${\{\hat{X_1}, \hat{X_2}, \hat{X_3}, \hat{X_4}, \hat{X_5}\}}$ with $\hat{X_k} \in \mathbb{R}^{H \times W \times C}$ of width $W$, height $H$, and $C$ channels) from the CSI module are the input of SA module. For each spatial position $(i,j)$, we concatenate the $K$ scale features to derive the attention over different scales and channels.

For each spatial position $(i,j)$, we represent the feature vectors from the five scale feature maps as $\mathbf{s}_k^{(i,j)} = \hat{X_k}(i,j) \in \mathbb{R}^C$, where $k = 1,\cdots,5$. We concatenate the feature vectors to have $\mathbf{s}^{(i,j)} = [\mathbf{s}_1^{(i,j)}, \mathbf{s}_2^{(i,j)}, \cdots, \mathbf{s}_5^{(i,j)}]$. Specifically, the channel adaptive scale attention vector $\mathbf{\alpha}_k^{(i,j)}$ over the $k^{th}$ refined feature map $\hat{X_k}$ at position $(i,j)$ is obtained by:
\begin{equation}
\begin{split}
    \mathbf{\alpha}_k^{(i,j)} = Sigmoid(W_k ReLU(W \mathbf{s}^{(i,j)})),
\end{split}
\end{equation}
where $W$ and $W_k$ are implemented by 1$\times$1 convolution followed by batch normalization. $W$ shrinks the channel dimension by a rate $r$ (we experimentally set it to $r=4$), and $W_k$ transforms the channel dimension to $C$. 

Then, the attention element-wise modulates the channel dimensions for each scale at position $(i,j)$:

\begin{equation}
    \begin{aligned}
        \mathbf{s}_{fusion}^{(i,j)} = \frac{1}{5}\sum_{k=1}^5 \mathbf{\alpha}_k^{(i,j)}\mathbf{s}_k^{(i,j)},
    \end{aligned}
    \label{eq:fusion}
\end{equation}
where $\mathbf{s}_{fusion}^{(i,j)}$ denotes the final feature that fuses the features at the same position $(i,j)$ from difference scales by scale attention weighting.



Similar idea but with shared channel attention has been proposed in \cite{chen2016attention,kong2019pixel} to spatial adaptively determine the suitable feature scales. However, they all constrain the sum of the attention weights across scales to be 1, \eg, using SoftMax activation function. We relax the sum 1 constraint on the attention weights by using $Sigmoid$ activation function. This provides more flexible optimization space. Intuitively, since the sum of the attention weights at a spatial position is not restricted to be 1, the energy at different spatial positions can vary and this is equivalent to playing a role of spatial attention. Further, to be more flexible, our scale attention is also channel adaptive.

\section{Experiments}

To evaluate the proposed CASINet, we carry out comprehensive experiments on three widely-used scene parsing datasets including Cityscapes \cite{cordts2016cityscapes}, ADE20K \cite{zhou2017scene} and LIP \cite{gong2017look}. In the following sections, we will first introduce three datasets and the implementation details, respectively. Then we perform extensive ablation studies on the Cityscapes dataset to demonstrate the effectiveness of our designs. Finally, we further evaluate CASINet's performance on ADE20K \cite{zhou2017scene} and LIP \cite{gong2017look}.

\subsection{Datasets}
\noindent\textbf{Cityscapes.} The dataset is a large, diverse set of high-resolution (1024$\times$2048) images recorded in streets from 50 different cities, where 5000 images have high quality pixel-level labels of 19 classes and are finely annotated. Following the standard setting of Cityscapes, the finely annotated 5000 images are divided into 2975, 500, and 1525 images for training, validation, and testing. We do not use coarse data (20000 coarsely annotated images) in our experiments. 

\noindent\textbf{ADE20K.} The dataset has been used in ImageNet scene parsing challenge 2016, including 150 classes and diverse scenes with 1038 image-level labels, which is divided into 20K/2K/3K images for training, validation, and testing, respectively.

\noindent\textbf{LIP.} The dataset has been used in the Look into Person (LIP) challenge 2016 for single human parsing, including 50,462 images with 19 semantic human part classes and 1 background class.

\subsection{Implementation Details}

We employ a pre-trained FCN (ResNet) \cite{wang2017residual} with the dilated strategy \cite{chen2018deeplab} as the backbone to extract the feature map. To capture objects of different scales, we organize five dilated convolutional branches in a parallel fashion, which provides five feature maps with each characterized by a different scale/receptive field. Referring to ASPP \cite{chen2018deeplab,chen2017rethinking}, we set the dilation rates for the five branches as $d = 1, 6, 12, 24, 36$, respectively. We implement our method based on Pytorch, and all experiments are performed on 4x Tesla P40 GPUs. 

\subsubsection{Cityscapes.} Referring to the public settings in previous work, we set the initial learning rate as 0.01 and weight decay as 0.0005. The original image size is 1024$\times$2048 and we crop it to 769$\times$769 following PSPNet \cite{zhao2017pyramid}. The training batch size is set to 8, and we train our model with InPlaceABNSync \cite{rota2018place} to synchronize the mean and standard-deviation of BN across multiple GPUs in all the experiments. We employ 40K/80k training iterations when training without/with validation set. Similar to the previous works ASPP \cite{chen2018deeplab} and DenseASPP \cite{yang2018denseaspp}, we also employ the poly learning rate policy where the learning rate is multiplied by ${(1-iter/total_{iter})^{0.9}}$. 

For the data augmentation, we apply random flipping horizontally and random scaling in the range of [0.5, 2]. For loss functions, following \cite{chen2018deeplab}, we employ class-balanced cross entropy loss on both the final output and the intermediate feature map output, where the weight of the final loss is 1 and the auxiliary loss is 0.4. Following previous works \cite{wu2016high}, we also adopt the online hard example mining (OHEM) strategy.

\subsubsection{ADE20K} We set the initial learning rate as 0.02 and weight decay as 0.0001. The input image is resized to {300, 375, 450, 525, 600} randomly since the images are of various sizes on ADE20K. The training batch size is set to 8 and we also train our model with InPlaceABNSync \cite{rota2018place}. We train the model for 100k iterations. By following the previous works \cite{chen2017rethinking,zhao2017pyramid}, we employ the same poly learning rate policy and data augmentation and employ supervision in the intermediate feature map.

\subsubsection{LIP} Following the CE2P \cite{liu2018devil}, we set the initial learning rate as 0.007 and weight decay as 0.0005. The original images are of various sizes and we resize all of them to 473$\times$473. The training batch size is set to 40 and we also train our model with InPlaceABNSync \cite{rota2018place}. We train the model for 110k iterations. The poly learning rate policy, data augmentation methods and deep supervision in the intermediate feature map are consistent with the experiments on Cityscapes and ADE20K.

\subsection{Ablation Study}
We perform all ablation studies on the Cityscapes datasets. We split this section into three sub-sections for better illustration: (1) CASINet versus Baseline. (2) Study on the Contextual Scale Interaction (CSI) module. (3) Study on the Scale Adaptation (SA) module.

\subsubsection{\textbf{CASINet vs. Baseline}} 

In order to verify the effectiveness of each module in CASINet, we perform comprehensive ablation studies on the Cityscapes validation datasets. We use the \textit{ResNet-101 Baseline} \cite{he2016deep} to represent the traditional FCN with the dilated strategy \cite{chen2018deeplab}, which can be viewed as the ``FCN'' in our pipeline (Figure~\ref{fig:pipeline}). To ensure  fairness, the \textit{ResNet-101 + ASPP} scheme follows DeepLabv3 \cite{chen2017rethinking} with some modifications in our designs: we remove the original image-level pooling branch, and employ five 3x3 dilated convolution branches with dilation rates of 1, 6, 12, 24 and 36, respectively (as shown in Figure~\ref{fig:pipeline}).

Because our key contributions lie in the contextual scale interaction (CSI) module and scale adaptation (SA) module, we verify their effectiveness separately: \textit{ResNet-101 + ASPP + CSI} and \textit{ResNet-101 + ASPP + SA}. The complete version, i.e. \textit{ResNet-101 + ASPP + CSI + SA}, is abbreviated as \textit{CASINet}. For quantitative evaluation, mean of class-wise Intersection over Union (mIoU) is used.

The experimental results are reported in Table \ref{table:ab}, where all the results are based on single scale testing. The performance of \textit{ResNet-101 + ASPP} is comparable to the numbers in the original DeepLabv3 paper \cite{chen2017rethinking}. We make the following observations.

\begin{table}[h]
\centering
\scriptsize
  \caption{Ablation study on Cityscapes validation set. CSI represents the contextual scale interaction (CSI) module, SA represents the scale adaptation (SA) module, and CASINet represents the complete version with both CSI and SA.}
  \label{table:ab}
\begin{tabular}{@{}lcc@{}}
\toprule
Methods                & Train. mIoU (\%) & Val. mIoU (\%) \\ \midrule
ResNet-101 Baseline    & 83.75            & 74.82          \\
ResNet-101 + ASPP \cite{chen2017rethinking}     & 85.91            & 78.51          \\ \midrule
ResNet-101 + ASPP + CSI  & 87.74	      & 80.17          \\
ResNet-101 + ASPP + SA & 87.66            & 80.08          \\ \midrule
CASINet                 & \textbf{88.33}            & \textbf{81.04}          \\ \bottomrule
\end{tabular}
\end{table}

Our contextual scale interaction (CSI) module and scale adaptation (SA) module both significantly improve over other two powerful baselines. \textit{ResNet-101 + ASPP + CSI} outperforms the \textit{ResNet-101 Baseline} and \textit{ResNet-101 + ASPP} by \textbf{5.35\%} and \textbf{1.66\%} respectively in mIoU, which verifies the effectiveness of the feature interaction among scales to exploit scale contextual information. In addition, \textit{ResNet-101 + ASPP + SA} outperforms the \textit{ResNet-101 + ASPP} by \textbf{1.57\%} in mIoU, which verifies the effectiveness of the spatial and channel adaptive scale attention.

We find that we can further improve the performance by combining the CSI and SA modules together. For example, the complete version \textit{CASINet} achieves the best (\textbf{81.04\%}) on the validation set based on single scale testing and improves by \textbf{0.87\%} over \textit{ResNet-101 + ASPP + CSI} and \textbf{0.96\%} over \textit{ResNet-101 + ASPP + SA}.

\noindent\textbf{Complexity.} The model size of our \textit{CASINet} (with ResNet-101 as backbone) is not much larger than that of \emph{ResNet-101 Baseline} (49.82 M vs. 44.59 M).

\begin{figure}[t]
  \includegraphics[width=1.0\linewidth]{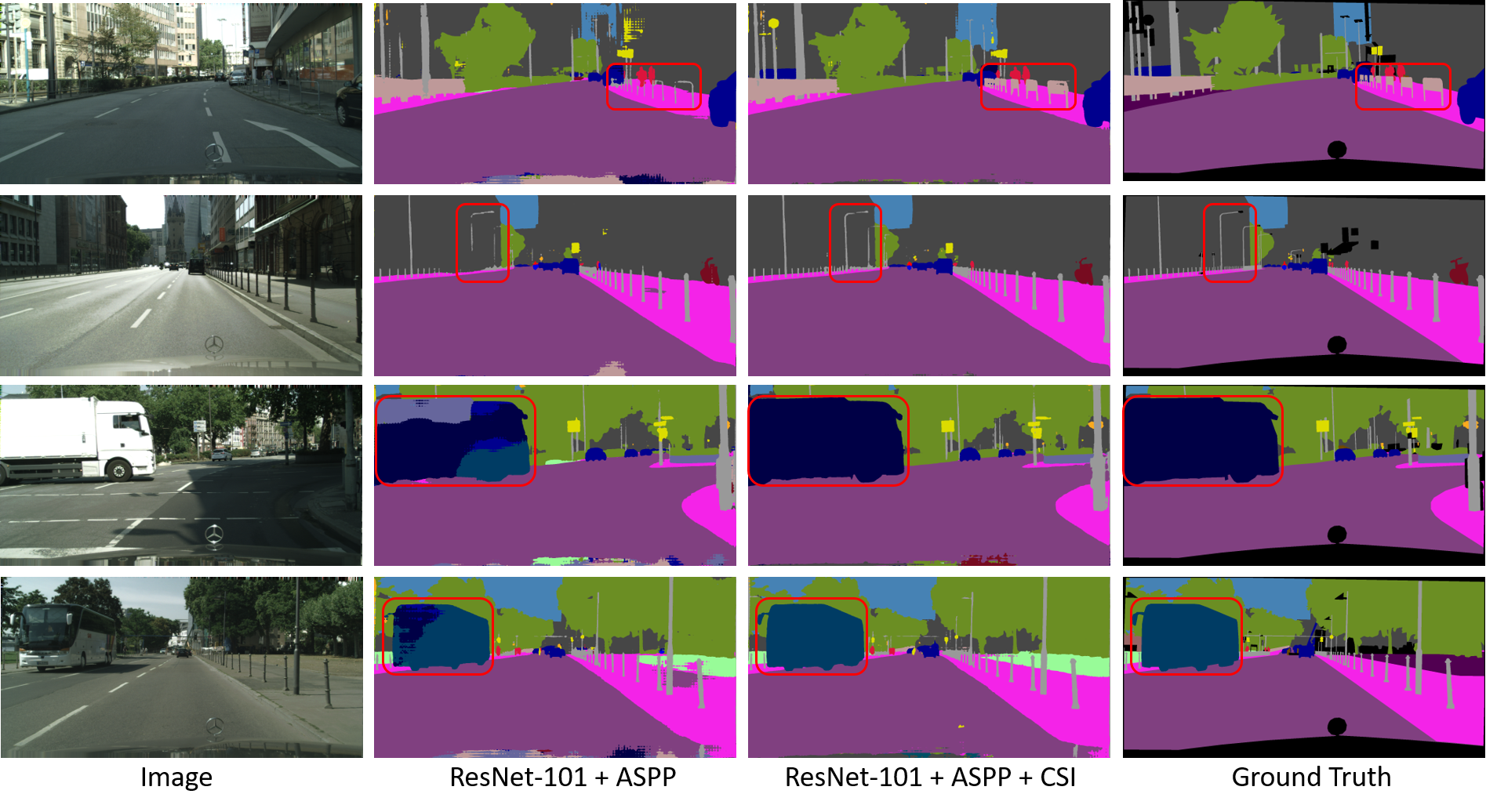}
  \caption{Visualization results of Contextual Scale Interaction (CSI) module on Cityscapes validation set.}
  \label{fig:RM_results}
\end{figure}

\subsubsection{\textbf{Study on Contextual Scale Interaction (CSI) Module}}

In this subsection, we study how different designs of the Contextual Scale Interaction (CSI) module influence performance quantitatively. 

\textbf{Influence of Scale Index.} The scale index (branch index with each corresponding to a different dilation rate) in our CASINet acts as an important scale prior to explicitly representing its corresponding receptive-field size. We verify its effectiveness in Table \ref{table:semantic}.

\begin{table}[th]
\vspace{-2mm}
\scriptsize
\centering
  \caption{Influence of the scale prior (scale index) in the CSI module evaluated on the validation set of Cityscapes. For fairness of comparison, we implement different designs on top of our baseline \textit{ResNet-101+ASPP}.}
  \label{table:semantic}
\tabcolsep=8.0pt  
\begin{tabular}{@{}lcc@{}}
\toprule
Methods                         & Train. mIoU (\%) & Val. mIoU (\%) \\ \midrule
CSI without scale index          & 86.68	        & 79.51         \\
CSI with scale index (Ours)      & \textbf{87.74}	        & \textbf{80.17}            \\ \bottomrule
\end{tabular}
\end{table}

\textbf{Sharing Embedding or Not?} We have compared the individual embedding and the shared embedding function among scales which are used for calculating the affinities within the CSI module. We find that the former achieves about \textbf{0.45\%} gain over the latter on the validation set. We note that features of different scales should be embedded into different high dimensional feature spaces. 

\begin{figure*}
  \centering
  \includegraphics[width=\linewidth]{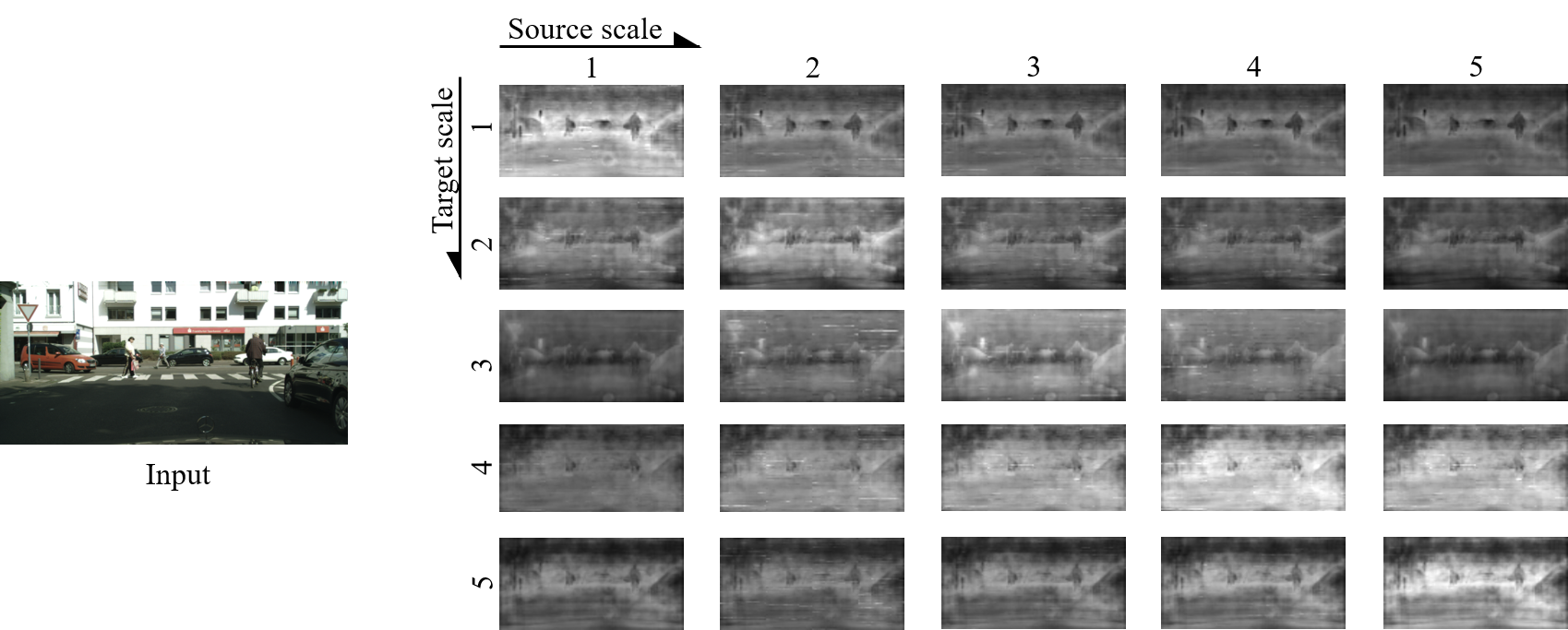}
  \vspace{-8mm}
  \caption{Visualization of the connecting weight (pairwise affinity) from the source scales (5 scales in total) to the target scale (5 scales in total) on all spatial positions. We randomly selected an image from the validation set of Cityscapes as the example.}
  \label{fig:AM}
\end{figure*}

\textbf{Visualization.} We visualize the influences of Contextual Scale Interaction (CSI) module on the parsing results in Figure~\ref{fig:RM_results}. We observe that some inconspicuous objects are well-captured and object boundaries are clearer with the CSI module included, such as the `fence' in the first row and the `pole' in the second row, which demonstrates that the contextual interaction over multi-scale features enhances the discrimination of details.
Meanwhile, some salient objects exhibit their intra-class consistency because some otherwise misclassified regions are now correctly classified with the CSI module, such as the `truck' in the third row and the `bus' in the fourth row, which further verifies that the feature interaction of CSI is powerful and can promote the content adaptation.

We further visualize the learned connecting weight $\tilde{c}_{k,m}^{(i,j)}$ as described around formulation (\ref{eq:refine}) from the source scales (5 scales in total) to the target scale (5 scales in total) on all spatial positions in Figure \ref{fig:AM}. We show them by 5$\times$5=25 gray images (5 columns denoting the 1-5 source scales while 5 rows denoting the 1-5 target scales respectively), with each pixel's value indicating the weight amplitudes. We observe that: (1) For different spatial positions, the learned weights/affinity values are different. They are spatially adaptive and thus can help to mimic the function of spatial adaptive filtering for better feature learning/refinement. (2) For different target scales (rows in Figure \ref{fig:AM}), the weight maps are different. In comparison with the largest scale 5 (the fifth row), the weights vary across the spatial positions at finer granularity and can better adapt to smaller objects. (3) For each target scale $k$, the weight map of the source scale $k$ (itself) has larger values than those of other scales, which indicates that the scale itself still dominates the feature representations in comparison with other source scales.

\subsubsection{\textbf{Study on Scale Adaptation (SA) Module}}

In this subsection, extensive experiments are designed to study how different designs of the Scale Adaptation (SA) module influence performance quantitatively. We perform the studies on the Cityscapes' validation set. For fairness of comparison, we implement different designs on top of our baseline \textit{ResNet-101 + ASPP}.

\textbf{Sigmoid or Softmax?} Our scale attention is different from the attention in \cite{chen2016attention,kong2019pixel} as they typically employ a Softmax/Gumbel Softmax function to map the un-normalized output to a probability distribution, which has to meet the sum 1 constraint on the attention weights across the candidates. We relax the sum 1 constraint on the attention weights by employing Sigmoid activation function across the scales. This provides more flexible optimization space. On the other hand, since the sum of the attention weights in a spatial position is not restricted to being 1, the energy at different spatial position can vary and play a role of spatial attention. We experimentally demonstrate the effectiveness of relaxing the sum 1 constraint and show the results in Table \ref{table:softmax}.

\begin{table}[ht]
\centering
\footnotesize
  \caption{Influence of relaxing the constraint on attention values in the SA module. We implement the different designs on top of our baseline \textit{ResNet-101+ASPP} for fair comparison.}
  \label{table:softmax}
\tabcolsep=10.8pt
\begin{tabular}{@{}lcc@{}}
\toprule
Methods                & Train. mIoU (\%) & Val. mIoU (\%) \\ \midrule
SA with Softmax        & 86.82               & 79.56             \\
SA with Sigmoid (Ours)        & \textbf{87.66}               & \textbf{80.08}             \\ \bottomrule
\end{tabular}
\end{table}

\begin{figure*}[th]
  \includegraphics[width=1.0\linewidth]{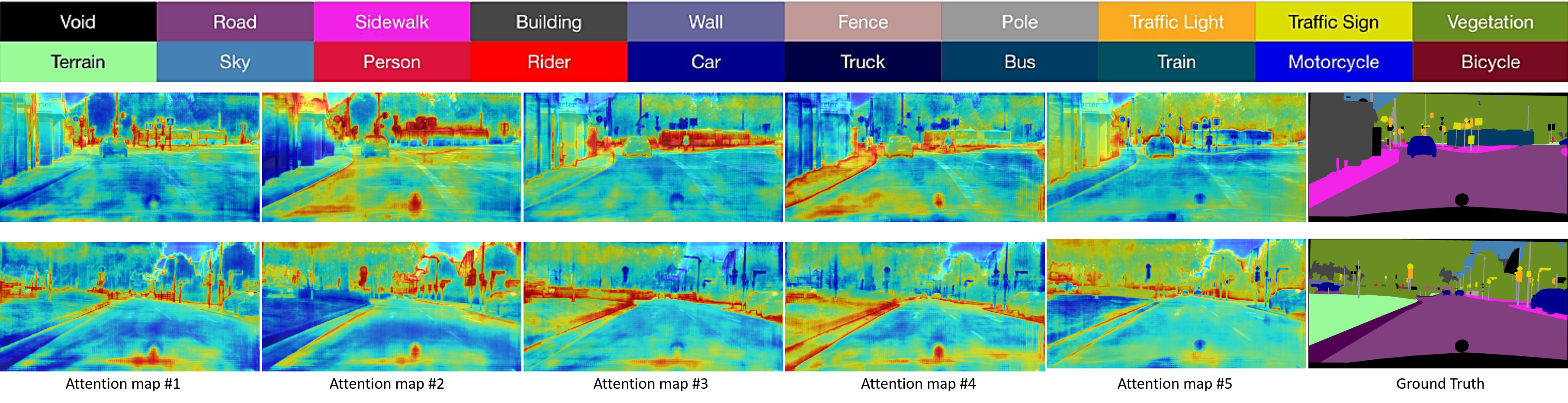}
  \vspace{-8mm}
  \caption{Visualization of scale attention (SA) map on five scales respectively for two images from Cityscapes validation set. Note that the attention maps typically have higher responses on the object edge, because the model has to put more energy on the boundary to achieve higher mIoU, so we can ignore this phenomenon and pay more attention to the difference between these maps.}
  \label{fig:Attmap}
\end{figure*}

\textbf{Attention Shared on Channels or Not?} For the scale adaptation, previous works \cite{chen2016attention,kong2019pixel} share the same attention value across channels for a scale and the attention values are different for different scales. To be more flexible, our attention is also channel adaptive. Table \ref{table:RA} shows the comparisons. \emph{ResNet-101 + ASPP + SA (shared)} denotes the scheme where attention value is shared across channels. \emph{ResNet-101 + ASPP + SA} denotes our scheme where attention is also adaptive on channels. Our SA achieves \textbf{0.58\%} gain in mIoU. Note that for different spatial positions, the attention responses are different and spatial adaptive. This is also illustrated in Figure~\ref{fig:Attmap}.
\begin{table}[th]
\footnotesize
\centering
  \caption{Comparisons of the Scale Adaptation (SA) with attention values shared across channels or unshared for a scale, evaluated on the validation set of Cityscapes. We implement on top of our baseline \textit{ResNet-101+ASPP} for fair comparison.} 
  \label{table:RA}
\begin{tabular}{@{}lcc@{}}
\toprule
Methods                    & Train. mIoU (\%) & Val. mIoU (\%) \\ \midrule
ResNet-101 + ASPP          & 85.91            & 78.51          \\
ResNet-101 + ASPP + SA (shared)     & 86.71            & 79.50           \\ \midrule
ResNet-101 + ASPP + SA   & \textbf{87.66}            & \textbf{80.08}          \\ \bottomrule
\end{tabular}
\vspace{-3mm}
\end{table}

\textbf{Visualization.} In order to further understand our Scale Adaptation (SA) module intuitively, we randomly choose some examples from the validation set of Cityscapes and visualize the Scale Attention (SA) map in Figure~\ref{fig:Attmap}. For each branch in CASINet, the overall scale attention map has a size of $H\times W \times C$ because it is spatial and channel adaptive, so we show the  scale attention map averaged along the channel dimension to see whether they could capture objects of different scales at different spatial positions.

As illustrated in Figure~ \ref{fig:Attmap}, we can find that each branch has different degrees of response to objects of different scales. For example, for the first and second branch with dilation rates being 1 and 6, their scale attention maps ``Attention map \#1'' and ``Attention map \#2'' mainly focus on small inconspicuous objects and object boundaries, such as the `traffic sign' in the first row and second row. With the increase of the dilation rates, the third and fourth branch with dilation rates being 12 and 24 both have larger receptive-fields, so their corresponding scale attention maps ``Attention map \#3'' and ``Attention map \#4'' mainly focus on salient objects, such as the `train, sidewalk' in the first row and the `terrain' in the second row. For the fifth (last) branch with dilation rates being 36, its receptive-field is large enough to cover the biggest objects/stuffs in the scene, so its ``Attention map \#5'' mainly focuses on the background region of the scene, such as the `building, vegetation' in the first row and second row. In short, these visualizations further verify that our Scale Adaptation (SA) module helps each branch focus on different objects with different scales, and adaptively select the appropriate receptive fields/scales for each spatial position.

\begin{figure*}[th]
  \subfigure{\includegraphics[width=1.0\linewidth]{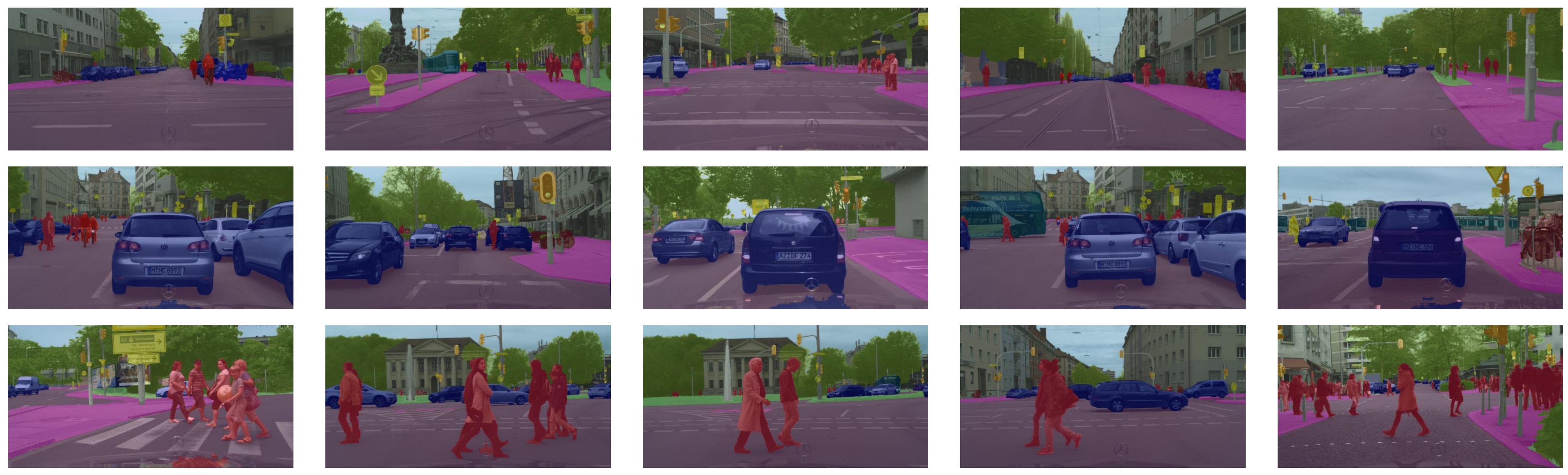}}
  \vspace{-8mm}
  \caption{Visualization of scene parsing results predicted by our CASINet from Cityscapes validation set. The images in the first row contain more large-scale objects, such as road. The second and the third row respectively present some middle-scale and small-scale objects, such as cars and people.}
  \label{fig:more_results}
\end{figure*}

To further demonstrate the effectiveness and superiority, we provide more segmentation results of our CASINet on the changeling Cityscapes dataset in Figure \ref{fig:more_results} where the images are randomly sampled from the validation set. The images in the first row contain more large-scale objects, such as road. The second and the third row respectively present some middle-scale and small-scale objects, such as cars and people. We observe that CASINet could obtain accurate parsing results for different objects with various scales, which indicates the effectiveness of our proposed CSI and SA modules.

\begin{table}[t]
\centering
  \caption{Influence of non-local operation on spatial,\ie, \textit{(+Non-local)}, or on scales, \ie, CSI. We implement the different designs on top of our scheme \textit{ResNet-101+ASPP+SA}. Being functionally different, the spatial-wise non-local and our scale-wise non-local are complementary to each other. }
  \label{table:non-local}
\resizebox{0.8\textwidth}{!}{  
\begin{tabular}{@{}lcc@{}}
\toprule
Methods                & Train. mIoU (\%) & Val. mIoU (\%) \\ \midrule
ResNet-101+ASPP+SA   & 87.66            & 80.08          \\ 
ResNet-101+ASPP+CSI+SA                 & \textbf{88.33}            & \textbf{81.04}          \\ \midrule

ResNet-101+ASPP(+Non-local)	& 87.28	& 79.62 \\

ResNet-101+ASPP(+Non-local)+SA        & 87.91               & 80.39             \\ 
ResNet-101+ASPP(+Non-local)+CSI+SA      & \textbf{88.68}          & \textbf{81.41}  \\
\bottomrule
\end{tabular}
}
\end{table}

\subsubsection{\textbf{Study on Non-local w.r.t. Spatial and Scales}}\label{not-non-local}
To further verify the difference and complementary relationship between the non-local operation on spatial \cite{wang2018non,yuan2018ocnet}, and ours on scales which enables interaction among scales in CSI module, we design another two schemes for comparisons as shown in Table \ref{table:non-local}. (1) \textit{ResNet-101+ASPP(+Non-local)+SA}: Based on the scheme \textit{ResNet-101+ASPP+SA}, we add an additional spatial-wise non-local branch (parallel to the five scale branches) similar to that in OCNet\cite{yuan2018ocnet}, before our SA module. (2) \textit{ResNet-101+ASPP(+Non-local)+CSI+SA}: Based on  \textit{ResNet-101+ASPP(+Non-local)+SA}, we add our CSI module for the scale interaction achieved by non-local operation across scales. We can observe that our CSI module (in scheme \textit{ResNet-101+ASPP(+Non-local)+CSI+SA}) brings \textbf{1.02\%} gain in Val. mIoU compared with the scheme \textit{ResNet-101+ASPP(+Non-local)+SA}. This indicates our CSI module is complementary to the spatial non-local idea. In comparison with the scheme \textit{ResNet-101+ASPP+SA}, the gain of our CSI (of scheme \textit{ResNet-101+ASPP+CSI+SA}) is \textbf{0.96\%}, which is larger than that of using spatial non-local (of scheme \textit{ResNet-101+ASPP(+Non-local)+SA}), \ie, \textbf{0.31\%}. This demonstrates the feature interactions among scales can enhance the power of feature representation.

\setlength{\tabcolsep}{1.8pt}
\begin{table*}[t]
\tiny
\centering
  \caption{Category-wise comparisons with state-of-the-art approaches on the Cityscapes test set. CASINet outperforms existing methods and achieves 81.9\% in Mean IoU. \textit{Baseline$^{1}$} denotes the scheme of \textit{ResNet-101+ASPP}. \textit{Baseline$^{2}$} denotes the scheme of  \textit{ResNet-101+ASPP(+Non-local)}. \textit{CASINet$^{1}$} denotes the scheme of \textit{ResNet-101+ASPP+CSI+SA}. \textit{CASINet$^{2}$} denotes the scheme of \textit{ResNet-101+ASPP(+Non-local)+CSI +SA}. $*$: Models are not pretrained on ImageNet.}
  \resizebox{1.0\textwidth}{!}{
  \label{table:STO}
  \scalebox{0.99}{
\begin{tabular}{@{}l|l|lllllllllllllllllll@{}}
\toprule
Methods    & \rotatebox{90}{Mean IoU} & \rotatebox{90}{road} & \rotatebox{90}{sidewalk} & \rotatebox{90}{building} & \rotatebox{90}{wall} & \rotatebox{90}{fence} & \rotatebox{90}{pole} & \rotatebox{90}{traffic light} & \rotatebox{90}{traffic sign} & \rotatebox{90}{vegetation} & \rotatebox{90}{terrain} & \rotatebox{90}{sky}  & \rotatebox{90}{person} & \rotatebox{90}{rider} & \rotatebox{90}{car}  & \rotatebox{90}{truck} & \rotatebox{90}{bus}  & \rotatebox{90}{train} & \rotatebox{90}{motorcycle} & \rotatebox{90}{bicycle} \\ \midrule
FCN-8s \cite{long2015fully}     & 65.3     & 97.4 & 78.4     & 89.2     & 34.9 & 44.2  & 47.4 & 60.1          & 65           & 91.4       & 69.3    & 93.9 & 77.1   & 51.4  & 92.6 & 35.3  & 48.6 & 46.5  & 51.6       & 66.8    \\
DeepLab-v2 \cite{chen2018deeplab} & 70.4     & 97.9 & 81.3     & 90.3     & 48.8 & 47.4  & 49.6 & 57.9          & 67.3         & 91.9       & 69.4    & 94.2 & 79.8   & 59.8  & 93.7 & 56.5  & 67.5 & 57.5  & 57.7       & 68.8    \\
FRRN \cite{pohlen2017full}       & 71.8     & 98.2 & 83.3     & 91.6     & 45.8 & 51.1  & 62.2 & 69.4          & 72.4         & 92.6       & 70      & 94.9 & 81.6   & 62.7  & 94.6 & 49.1  & 67.1 & 55.3  & 53.5       & 69.5    \\
RefineNet \cite{lin2017refinenet}  & 73.6     & 98.2 & 83.3     & 91.3     & 47.8 & 50.4  & 56.1 & 66.9          & 71.3         & 92.3       & 70.3    & 94.8 & 80.9   & 63.3  & 94.5 & 64.6  & 76.1 & 64.3  & 62.2       & 70      \\
GCN \cite{peng2017large}       & 76.9     & -    & -        & -        & -    & -     & -    & -             & -            & -          & -       & -    & -      & -     & -    & -     & -    & -     & -          & -       \\
DUC \cite{wang2018understanding}       & 77.6     & 98.5 & 85.5     & 92.8     & 58.6 & 55.5  & 65   & 73.5          & 77.9         & 93.3       & 72      & 95.2 & 84.8   & 68.5  & 95.4 & 70.9  & 78.8 & 68.7  & 65.9       & 73.8    \\
ResNet-38 \cite{wu2019wider} & 78.4     & 98.5 & 85.7     & 93.1     & 55.5 & 59.1  & 67.1 & 74.8          & 78.7         & 93.7       & 72.6    & 95.5 & 86.6   & 69.2  & 95.7 & 64.5  & 78.8 & 74.1  & 69         & 76.7    \\
DSSPN \cite{liang2018dynamic}    & 77.8     & -    & -        & -        & -    & -     & -    & -             & -            & -          & -       & -    & -      & -     & -    & -     & -    & -     & -          & -       \\
DepthSeg \cite{kong2018recurrent} & 78.2    & -    & -        & -        & -    & -     & -    & -             & -            & -          & -       & -    & -      & -     & -    & -     & -    & -     & -          & -       \\
PSPNet \cite{zhao2017pyramid}    & 78.4     & -    & -        & -        & -    & -     & -    & -             & -            & -          & -       & -    & -      & -     & -    & -     & -    & -     & -          & -       \\
BiSeNet \cite{yu2018bisenet}   & 78.9     & -    & -        & -        & -    & -     & -    & -             & -            & -          & -       & -    & -      & -     & -    & -     & -    & -     & -          & -       \\
CiSS-Net \cite{zhou2019context} & 79.2     & -    & -        & -        & -    & -     & -    & -             & -            & -          & -       & -    & -      & -     & -    & -     & -    & -     & -          & -       \\
DFN \cite{yu2018learning}   & 79.3      & -    & -        & -        & -    & -     & -    & -             & -            & -          & -       & -    & -      & -     & -    & -     & -    & -     & -          & -       \\
PSANet \cite{zhao2018psanet}    & 80.1     & -    & -        & -        & -    & -     & -    & -             & -            & -          & -       & -    & -      & -     & -    & -     & -    & -     & -          & -       \\
Auto-DeepLab-S* \cite{liu2019auto} & 79.9     & -    & -        & -        & -    & -     & -    & -             & -            & -          & -       & -    & -      & -     & -    & -     & -    & -     & -          & -       \\

Auto-DeepLab-L* \cite{liu2019auto} & 80.4     & -    & -        & -        & -    & -     & -    & -             & -            & -          & -       & -    & -      & -     & -    & -     & -    & -     & -          & -       \\
DenseASPP \cite{yang2018denseaspp} & 80.6     & \textbf{98.7} & 87.1     & 93.4     & 60.7 & 62.7  & 65.6 & 74.6          & 78.5         & 93.6       & 72.5    & 95.4 & 86.2   & 71.9  & 96.0 & 78.0  & 90.3 & 80.7  & 69.7       & 76.8    \\
DANet \cite{fu2019dual} &81.5 &98.6 &86.1 &93.5 &56.1 &63.3 &\textbf{69.7} &77.3 &\textbf{81.3} &93.9 &72.9 &95.7 &\textbf{87.3} &72.9 &96.2 &76.8 &89.4 &86.5 &\textbf{72.2} &\textbf{78.2} \\
OCNet \cite{yuan2018ocnet} & 81.7 & -    & -        & -        & -    & -     & -    & -             & -            & -          & -       & -    & -      & -     & -    & -     & -    & -     & -          & -       \\
GALD \cite{li2019global} & 81.8 & -    & -        & -        & -    & -     & -    & -             & -            & -          & -       & -    & -      & -     & -    & -     & -    & -     & -          & -       \\
ACFNet \cite{zhang2019acfnet} & 81.8 & -    & -        & -        & -    & -     & -    & -             & -            & -          & -       & -    & -      & -     & -    & -     & -    & -     & -          & -       \\
DGCNet \cite{zhang2019dual} & 82.0 & -    & -        & -        & -    & -     & -    & -             & -            & -          & -       & -    & -      & -     & -    & -     & -    & -     & -          & -       \\

\midrule
Baseline$^{1}$     & 78.1     & 98.6    & 86.2        & 93.0        & 57.7    & 60.7     & 64.7    & 73.7             & 77.6            & 93.4          & 72.9       & 95.2    & 85.5      & 70.1     & 95.8    & 66.8     & 81.7    & 66.0     & 68.3          & 75.6          \\ 
CASINet$^{1}$     & {81.9}     & \textbf{98.7} & {87.2}     & \textbf{93.7}     & \textbf{62.6} & \textbf{64.7}  & {69.0}   & {76.4}          & {80.7}         & {93.7}       & {73.3}    & {95.6} & {86.8}   & {72.3}  & {96.2} & {78.1}  & {90.6} & \textbf{87.9}  & {71.5}       & {76.9}   \\  \midrule
Baseline$^{2}$     & 79.5     & \textbf{98.7}       & 86.9        & 93.4        & 59.4    & 61.6     & 67.2    & 75.3             & 79.0            & 93.6          & 73.2       & 95.4    & 86.4      & 71.3     & 95.9    & 70.1     & 84.6    & 74.1     & 69.3          & 76.5          \\ 
CASINet$^{2}$     & \textbf{82.2}     & \textbf{98.7} & \textbf{87.3}     & {93.6}     & {62.0} & {62.9}  & {69.5}   & \textbf{77.8}          & {81.0}         & \textbf{94.1}       & \textbf{73.4}    & \textbf{95.8} & {87.1}   & \textbf{73.5}  & \textbf{96.3} & \textbf{80.1}  & \textbf{91.3} & {87.6}  & {72.0}       & {78.0}          \\ \bottomrule
\end{tabular}}}
\end{table*}

\subsection{Comparison with State-of-the-Art}

\textbf{Results on Cityscapes Dataset.} We compare our method with existing methods on the Cityscapes test set. Specifically, we train our CASINet with only finely annotated data (including validation set for training) for 80k training iterations and submit our test results (with multi-scale [0.75x, 1.0x, 1.25x] testing strategy) to the official evaluation server. Results are shown in Table \ref{table:STO}. \emph{Baseline$^{1}$} denotes the scheme of \emph{ResNet-101+ASPP}. \emph{Baseline$^{2}$} denotes the scheme of \emph{ResNet-101+ASPP(+Non-local)}. \emph{CASINet$^{1}$} denotes the scheme of \emph{ResNet-101+ASPP+CSI+SA}. \emph{CASINet$^{2}$} denotes the scheme of \emph{ResNet-101+\\ASPP(+Non-local)+CSI+SA}. We observe that \emph{CASINet} outperforms existing approaches. In particular, our \emph{CASINet$^{1}$} outperforms \emph{DenseASPP} \cite{yang2018denseaspp} by \textbf{1.3\%} in mIoU, even though \emph{DenseASPP} uses a more powerful and more complex pretrained DenseNet model \cite{huang2017densely} as the backbone network. In addition, compared with two baseline schemes \emph{Baseline$^{1}$}, and \emph{Baseline$^{2}$}, our two schemes \emph{CASINet$^{1}$} and \emph{CASINet$^{2}$} obtain \textbf{3.8\%} and \textbf{2.7\%} gains in mIoU, respectively, which  demonstrates the effectiveness and robustness of our proposed CSI and SA modules. Compared with the approaches of OCNet \cite{yuan2018ocnet} and DANet \cite{fu2019dual}, our CASINet$^{2}$ also achieves superior performance. 


\textbf{Results on ADE20K Dataset.} We further carry out experiments on the ADE20K dataset to evaluate the effectiveness of our method. Comparisons with previous state-of-the-art methods on the ADE20K validation set are shown in Table \ref{table:ADE}. Results show that our CASINet achieves the best performance of 45.61\% in Mean IoU. Our CASINet$^1$ outperforms the ResNet-101 based state-of-the-art method EncNet by \textbf{0.63\%} in mIoU, and improves over the ResNet-269 based PSPNet by \textbf{0.34\%} in mIoU. Note that we have not tuned the parameters of our networks (baselines and ours) aggressively. The performance of our \emph{Baseline$^2$} (that also adds a spatial-wise non-local branch as OCNet) is 1.78\% lower than OCNet\cite{yuan2018ocnet} in mIOU. The introduction of our modules improves  \emph{Baseline$^2$} significantly by 1.94\% in mIOU and achieves the best performance (\textbf{45.61\%} in mIoU).

\begin{table}[http]
\centering
\renewcommand\arraystretch{1.5}
\scriptsize
  \caption{Comparisons with state-of-the-art approaches on the validation set (to be consistent with what were done in prior works for performance comparisons) of the ADE20K dataset. \textit{Baseline$^{1}$} denotes the scheme of \emph{ResNet-101+ASPP}. \textit{Baseline$^{2}$} denotes the scheme of \emph{ResNet-101+ASPP(+Non-local)}. \textit{CASINet$^{1}$} denotes the scheme of \emph{ResNet-101+ASPP+CSI+SA}. \emph{CASINet$^{2}$} denotes the scheme of \emph{ResNet-101+ASPP(+Non-local)+CSI+SA}. $*$: Models are not pretrained on ImageNet.}
  \label{table:ADE}
\tabcolsep=8.8pt  
\begin{tabular}{@{}lccc@{}}
\toprule
Methods   & Publication & Backbone   & Val. mIoU (\%) \\ \midrule
RefineNet \cite{lin2017refinenet} & CVPR2017    & ResNet-101 & 40.20          \\
RefineNet \cite{lin2017refinenet} & CVPR2017    & ResNet-152 & 40.70          \\
PSPNet \cite{zhao2017pyramid}   & CVPR2017    & ResNet-101 & 43.29          \\
PSPNet \cite{zhao2017pyramid}   & CVPR2017    & ResNet-152 & 43.51          \\
PSPNet \cite{zhao2017pyramid}   & CVPR2017    & ResNet-269 & 44.94          \\
SAC \cite{zhang2017scale}      & ICCV2017    & ResNet-101 & 44.30          \\
PSANet \cite{zhao2018psanet}   & ECCV2018    & ResNet-101 & 43.77          \\
UperNet \cite{xiao2018unified}  & ECCV2018    & ResNet-101 & 42.66          \\
DSSPN \cite{liang2018dynamic}    & CVPR2018    & ResNet-101 & 43.68          \\
EncNet \cite{zhang2018context}   & CVPR2018    & ResNet-101 & 44.65          \\
OCNet \cite{yuan2018ocnet} & ArXiv2018    & ResNet-101 & 45.45          \\
CiSS-Net \cite{zhou2019context}  & CVPR2019 & ResNet-50 & 42.56         \\
CFNet \cite{zhang2019co} & CVPR2019 & ResNet-50 & 42.87         \\
CFNet \cite{zhang2019co} & CVPR2019 & ResNet-101 & 44.89         \\
Auto-DeepLab-S* \cite{liu2019auto} & CVPR2019 & -- & 40.69         \\
Auto-DeepLab-M* \cite{liu2019auto} & CVPR2019 & -- & 42.19         \\
Auto-DeepLab-L* \cite{liu2019auto} & CVPR2019 & -- & 43.98         \\
APCNet \cite{he2019adaptive} & CVPR2019 & ResNet-101 & 45.38         \\
\midrule
Baseline$^{1}$    & --          & ResNet-101 & {42.88}          \\ 
CASINet$^{1}$    & --          & ResNet-101 & {45.28}          \\ \midrule
Baseline$^{2}$    & --          & ResNet-101 & {43.67}          \\ 
CASINet$^{2}$    & --          & ResNet-101 & \textbf{45.61}          \\
\bottomrule
\end{tabular}
\end{table}

\textbf{Results on LIP Dataset.} We also conduct experiments on the LIP dataset. Comparisons with  previous state-of-the-art methods are reported in Table \ref{table:LIP}. We observe that the proposed \emph{CASINet$^2$} achieves \textbf{55.14\%} in mIoU, which outperforms \emph{Baseline$^2$} by \textbf{3.73\%} thanks to the introduction of our modules, indicating the effectiveness of our modules on the human parsing task.

Furthermore, recent released work \cite{wang2019learning,wang2020hierarchical} both propose to combine neural networks with the compositional fine-grained hierarchy of human bodies for complete human parsing and achieves 57.74\%, 59.25\% in mean IoU. But, the two methods both employ heavy multi-level semantic parts of human body, while our method is totally semantic free and thus is potentially been further improved by additionally leveraging such information” and add corresponding results in our Table \ref{table:LIP}.

\begin{table}[th]
\scriptsize
\centering
  \caption{Comparisons with state-of-the-art approaches on the validation set of the LIP dataset. \textit{Baseline$^{1}$} denotes the scheme of \textit{ResNet-101+ASPP}. \textit{Baseline$^{2}$} denotes the scheme of \textit{ResNet-101+ASPP(+Non-local)}. \textit{CASINet$^{1}$} denotes the scheme of \textit{ResNet-101+ASPP+CSI+SA}. \textit{CASINet$^{2}$} denotes the scheme of \textit{ResNet-101+ASPP(+Non-local)+CSI+SA}.}
  \label{table:LIP}
\tabcolsep=5.8pt   
\begin{tabular}{@{}lccc@{}}
\toprule
Methods       & Publication & Backbone   & Val. mIoU (\%) \\ \midrule
Attention+SSL \cite{gong2017look} & CVPR2017    & ResNet-101 & 44.73          \\
JPPNet \cite{liang2018look}        & TPAMI2018   & ResNet-101 & 51.37          \\
SS-NAN \cite{zhao2017self}        & CVPR2017    & ResNet-101 & 47.92          \\
MMAN \cite{luo2018macro}         & ECCV2018    & ResNet-101 & 46.81          \\
MuLA \cite{nie2018mutual}         & ECCV2018    & ResNet-101 & 49.30          \\
CE2P \cite{liu2018devil}         & AAAI2019    & ResNet-101 & 53.10          \\
OCNet \cite{yuan2018ocnet} & ArXiv2018    & ResNet-101 & 54.72          \\
CNIF \cite{wang2019learning} & ICCV2019 & ResNet-101 & 57.74 \\
HHP \cite{wang2020hierarchical} & CVPR2020 & ResNet-101 & \textbf{59.25} \\
\midrule
Baseline$^{1}$    & --          & ResNet-101 & {49.59}          \\ 
CASINet$^{1}$    & --          & ResNet-101 & {54.38}          \\ \midrule
Baseline$^{2}$    & --          & ResNet-101 & {51.41}          \\ 
CASINet$^{2}$    & --          & ResNet-101 & {55.14}          \\ \bottomrule
\end{tabular}
\end{table}

\section{Conclusion}

In this paper, we propose a simple yet effective Content-Adaptive Scale Interaction Network (CASINet) to adaptively exploit multi-scale features through contextual interaction and adaptation. Specifically, we build the framework based on the classic Atrous Spatial Pyramid Pooling (ASPP), followed by the proposed contextual scale interaction (CSI) module which enables spatial adaptive filtering for better feature learning, and a spatial and channel adaptive scale adaptation (SA) module to facilitates the appropriate scale selection. Our ablation studies demonstrate the effectiveness of the proposed CSI and SA modules, leading to more precise parsing results. In addition, CASINet achieves state-of-the-art performance on Cityscapes, ADE20K, and LIP datasets.

Although extensive experiments have demonstrated the effectiveness of the proposed CSI and SA modules, our method still has some limitations and will fail in some cases. For example, when the occlusion situation and crowded-overlap problem existed in the parsing image, the scale interaction effect within CSI may hurt the discrimination capability in the junction of different semantic categories and thus drop the parsing performance. We will leave such problem as our future work and further improve our algorithm.

\bibliographystyle{elsarticle-num}
\bibliography{mybibfile}

\begin{thebibliography}{84}
\expandafter\ifx\csname natexlab\endcsname\relax\def\natexlab#1{#1}\fi
\providecommand{\url}[1]{\texttt{#1}}
\providecommand{\href}[2]{#2}
\providecommand{\path}[1]{#1}
\providecommand{\DOIprefix}{doi:}
\providecommand{\ArXivprefix}{arXiv:}
\providecommand{\URLprefix}{URL: }
\providecommand{\Pubmedprefix}{pmid:}
\providecommand{\doi}[1]{\href{http://dx.doi.org/#1}{\path{#1}}}
\providecommand{\Pubmed}[1]{\href{pmid:#1}{\path{#1}}}
\providecommand{\bibinfo}[2]{#2}
\ifx\xfnm\relax \def\xfnm[#1]{\unskip,\space#1}\fi
\bibitem[{Lateef and Ruichek(2019)}]{lateef2019survey}
\bibinfo{author}{F.~Lateef}, \bibinfo{author}{Y.~Ruichek},
\newblock \bibinfo{title}{Survey on semantic segmentation using deep learning
  techniques},
\newblock \bibinfo{journal}{Neurocomputing} \bibinfo{volume}{338}
  (\bibinfo{year}{2019}) \bibinfo{pages}{321--348}.
\bibitem[{Wang et~al.(2019)Wang, Li, Aertsen, Deprest, Ourselin, and
  Vercauteren}]{wang2019aleatoric}
\bibinfo{author}{G.~Wang}, \bibinfo{author}{W.~Li},
  \bibinfo{author}{M.~Aertsen}, \bibinfo{author}{J.~Deprest},
  \bibinfo{author}{S.~Ourselin}, \bibinfo{author}{T.~Vercauteren},
\newblock \bibinfo{title}{Aleatoric uncertainty estimation with test-time
  augmentation for medical image segmentation with convolutional neural
  networks},
\newblock \bibinfo{journal}{Neurocomputing} \bibinfo{volume}{338}
  (\bibinfo{year}{2019}) \bibinfo{pages}{34--45}.
\bibitem[{Li et~al.(2020)Li, Liu, Liu, and Guo}]{li2020weakly}
\bibinfo{author}{Y.~Li}, \bibinfo{author}{Y.~Liu}, \bibinfo{author}{G.~Liu},
  \bibinfo{author}{M.~Guo},
\newblock \bibinfo{title}{Weakly supervised semantic segmentation by iterative
  superpixel-crf refinement with initial clues guiding},
\newblock \bibinfo{journal}{Neurocomputing}  (\bibinfo{year}{2020}).
\bibitem[{Zhang et~al.(2020)Zhang, Li, Hui, Liu, and Guan}]{zhang2020mfenet}
\bibinfo{author}{B.~Zhang}, \bibinfo{author}{W.~Li}, \bibinfo{author}{Y.~Hui},
  \bibinfo{author}{J.~Liu}, \bibinfo{author}{Y.~Guan},
\newblock \bibinfo{title}{Mfenet: Multi-level feature enhancement network for
  real-time semantic segmentation},
\newblock \bibinfo{journal}{Neurocomputing}  (\bibinfo{year}{2020}).
\bibitem[{Sun et~al.(2019)Sun, Zhao, Jiang, Cheng, Xiao, Liu, Mu, Wang, Liu,
  and Wang}]{sun2019high}
\bibinfo{author}{K.~Sun}, \bibinfo{author}{Y.~Zhao},
  \bibinfo{author}{B.~Jiang}, \bibinfo{author}{T.~Cheng},
  \bibinfo{author}{B.~Xiao}, \bibinfo{author}{D.~Liu}, \bibinfo{author}{Y.~Mu},
  \bibinfo{author}{X.~Wang}, \bibinfo{author}{W.~Liu},
  \bibinfo{author}{J.~Wang},
\newblock \bibinfo{title}{High-resolution representations for labeling pixels
  and regions},
\newblock \bibinfo{journal}{arXiv preprint arXiv:1904.04514}
  (\bibinfo{year}{2019}).
\bibitem[{Li et~al.(2019)Li, Zhang, You, Yang, Yang, and Tong}]{li2019global}
\bibinfo{author}{X.~Li}, \bibinfo{author}{L.~Zhang}, \bibinfo{author}{A.~You},
  \bibinfo{author}{M.~Yang}, \bibinfo{author}{K.~Yang},
  \bibinfo{author}{Y.~Tong},
\newblock \bibinfo{title}{Global aggregation then local distribution in fully
  convolutional networks},
\newblock \bibinfo{journal}{BMVC}  (\bibinfo{year}{2019}).
\bibitem[{Tian and Lin(2016)}]{tian2016video}
\bibinfo{author}{Z.~Tian}, \bibinfo{author}{Y.~Lin},
\newblock \bibinfo{title}{Video object segmentation based on supervoxel for
  multimedia corpus construction},
\newblock \bibinfo{journal}{Neurocomputing} \bibinfo{volume}{215}
  (\bibinfo{year}{2016}) \bibinfo{pages}{128--137}.
\bibitem[{Wang et~al.(2016)Wang, Fu, Xiong, and Zheng}]{wang2016transductive}
\bibinfo{author}{B.~Wang}, \bibinfo{author}{Z.~Fu}, \bibinfo{author}{H.~Xiong},
  \bibinfo{author}{Y.~F. Zheng},
\newblock \bibinfo{title}{Transductive video segmentation on tree-structured
  model},
\newblock \bibinfo{journal}{IEEE TCSVT} \bibinfo{volume}{27}
  (\bibinfo{year}{2016}) \bibinfo{pages}{992--1005}.
\bibitem[{Sun and Lu(2016)}]{sun2016interactive}
\bibinfo{author}{C.~Sun}, \bibinfo{author}{H.~Lu},
\newblock \bibinfo{title}{Interactive video segmentation via local appearance
  model},
\newblock \bibinfo{journal}{IEEE TCSVT} \bibinfo{volume}{27}
  (\bibinfo{year}{2016}) \bibinfo{pages}{1491--1501}.
\bibitem[{Zhu et~al.(2019)Zhu, Sapra, Reda, Shih, Newsam, Tao, and
  Catanzaro}]{zhu2019improving}
\bibinfo{author}{Y.~Zhu}, \bibinfo{author}{K.~Sapra}, \bibinfo{author}{F.~A.
  Reda}, \bibinfo{author}{K.~J. Shih}, \bibinfo{author}{S.~Newsam},
  \bibinfo{author}{A.~Tao}, \bibinfo{author}{B.~Catanzaro},
\newblock \bibinfo{title}{Improving semantic segmentation via video propagation
  and label relaxation},
\newblock in: \bibinfo{booktitle}{CVPR}, \bibinfo{year}{2019}, pp.
  \bibinfo{pages}{8856--8865}.
\bibitem[{Zhang et~al.(2019)Zhang, Lu, Lu, and Zhou}]{zhang2019refined}
\bibinfo{author}{L.~Zhang}, \bibinfo{author}{Y.~Lu}, \bibinfo{author}{L.~Lu},
  \bibinfo{author}{T.~Zhou},
\newblock \bibinfo{title}{Refined video segmentation through global appearance
  regression},
\newblock \bibinfo{journal}{Neurocomputing} \bibinfo{volume}{334}
  (\bibinfo{year}{2019}) \bibinfo{pages}{59--67}.
\bibitem[{Wang et~al.(2020)Wang, Ma, and You}]{wang2020deep}
\bibinfo{author}{X.~Wang}, \bibinfo{author}{H.~Ma}, \bibinfo{author}{S.~You},
\newblock \bibinfo{title}{Deep clustering for weakly-supervised semantic
  segmentation in autonomous driving scenes},
\newblock \bibinfo{journal}{Neurocomputing} \bibinfo{volume}{381}
  (\bibinfo{year}{2020}) \bibinfo{pages}{20--28}.
\bibitem[{Zhang et~al.(2018)Zhang, Chen, He, Ye, Cai, and
  Zhang}]{zhang2018road}
\bibinfo{author}{Y.~Zhang}, \bibinfo{author}{H.~Chen}, \bibinfo{author}{Y.~He},
  \bibinfo{author}{M.~Ye}, \bibinfo{author}{X.~Cai},
  \bibinfo{author}{D.~Zhang},
\newblock \bibinfo{title}{Road segmentation for all-day outdoor robot
  navigation},
\newblock \bibinfo{journal}{Neurocomputing} \bibinfo{volume}{314}
  (\bibinfo{year}{2018}) \bibinfo{pages}{316--325}.
\bibitem[{Park et~al.(2018)Park, Yang, Ro, Byun, Chae, and Han}]{park2018meet}
\bibinfo{author}{Y.~J. Park}, \bibinfo{author}{Y.~Yang},
  \bibinfo{author}{H.~Ro}, \bibinfo{author}{J.~Byun},
  \bibinfo{author}{S.~Chae}, \bibinfo{author}{T.~D. Han},
\newblock \bibinfo{title}{Meet ar-bot: Meeting anywhere, anytime with movable
  spatial ar robot},
\newblock in: \bibinfo{booktitle}{2018 ACM Multimedia},
  \bibinfo{organization}{ACM}, \bibinfo{year}{2018}, pp.
  \bibinfo{pages}{1242--1243}.
\bibitem[{Xie et~al.(2019)Xie, Huang, Jin, Liu, Zhu, Gao, and
  Zhang}]{xie2019weakly}
\bibinfo{author}{Z.~Xie}, \bibinfo{author}{Y.~Huang}, \bibinfo{author}{L.~Jin},
  \bibinfo{author}{Y.~Liu}, \bibinfo{author}{Y.~Zhu}, \bibinfo{author}{L.~Gao},
  \bibinfo{author}{X.~Zhang},
\newblock \bibinfo{title}{Weakly supervised precise segmentation for historical
  document images},
\newblock \bibinfo{journal}{Neurocomputing} \bibinfo{volume}{350}
  (\bibinfo{year}{2019}) \bibinfo{pages}{271--281}.
\bibitem[{Zhao et~al.(2019)Zhao, Chang, and Guo}]{zhao2019multimodal}
\bibinfo{author}{D.~Zhao}, \bibinfo{author}{Z.~Chang},
  \bibinfo{author}{S.~Guo},
\newblock \bibinfo{title}{A multimodal fusion approach for image captioning},
\newblock \bibinfo{journal}{Neurocomputing} \bibinfo{volume}{329}
  (\bibinfo{year}{2019}) \bibinfo{pages}{476--485}.
\bibitem[{Long et~al.(2015)Long, Shelhamer, and Darrell}]{long2015fully}
\bibinfo{author}{J.~Long}, \bibinfo{author}{E.~Shelhamer},
  \bibinfo{author}{T.~Darrell},
\newblock \bibinfo{title}{Fully convolutional networks for semantic
  segmentation},
\newblock in: \bibinfo{booktitle}{CVPR}, \bibinfo{year}{2015}, pp.
  \bibinfo{pages}{3431--3440}.
\bibitem[{Zhang et~al.(2017)Zhang, Tang, Zhang, Li, and Yan}]{zhang2017scale}
\bibinfo{author}{R.~Zhang}, \bibinfo{author}{S.~Tang},
  \bibinfo{author}{Y.~Zhang}, \bibinfo{author}{J.~Li},
  \bibinfo{author}{S.~Yan},
\newblock \bibinfo{title}{Scale-adaptive convolutions for scene parsing},
\newblock in: \bibinfo{booktitle}{ICCV}, \bibinfo{year}{2017}, pp.
  \bibinfo{pages}{2031--2039}.
\bibitem[{Zhao et~al.(2017)Zhao, Shi, Qi, Wang, and Jia}]{zhao2017pyramid}
\bibinfo{author}{H.~Zhao}, \bibinfo{author}{J.~Shi}, \bibinfo{author}{X.~Qi},
  \bibinfo{author}{X.~Wang}, \bibinfo{author}{J.~Jia},
\newblock \bibinfo{title}{Pyramid scene parsing network},
\newblock in: \bibinfo{booktitle}{CVPR}, \bibinfo{year}{2017}, pp.
  \bibinfo{pages}{2881--2890}.
\bibitem[{Chen et~al.(2018)Chen, Papandreou, Kokkinos, Murphy, and
  Yuille}]{chen2018deeplab}
\bibinfo{author}{L.-C. Chen}, \bibinfo{author}{G.~Papandreou},
  \bibinfo{author}{I.~Kokkinos}, \bibinfo{author}{K.~Murphy},
  \bibinfo{author}{A.~L. Yuille},
\newblock \bibinfo{title}{Deeplab: Semantic image segmentation with deep
  convolutional nets, atrous convolution, and fully connected crfs},
\newblock \bibinfo{journal}{IEEE TPAMI} \bibinfo{volume}{40}
  (\bibinfo{year}{2018}) \bibinfo{pages}{834--848}.
\bibitem[{Yang et~al.(2018)Yang, Yu, Zhang, Li, and Yang}]{yang2018denseaspp}
\bibinfo{author}{M.~Yang}, \bibinfo{author}{K.~Yu}, \bibinfo{author}{C.~Zhang},
  \bibinfo{author}{Z.~Li}, \bibinfo{author}{K.~Yang},
\newblock \bibinfo{title}{Denseaspp for semantic segmentation in street
  scenes},
\newblock in: \bibinfo{booktitle}{CVPR}, \bibinfo{year}{2018}, pp.
  \bibinfo{pages}{3684--3692}.
\bibitem[{Chen et~al.(2017)Chen, Papandreou, Schroff, and
  Adam}]{chen2017rethinking}
\bibinfo{author}{L.-C. Chen}, \bibinfo{author}{G.~Papandreou},
  \bibinfo{author}{F.~Schroff}, \bibinfo{author}{H.~Adam},
\newblock \bibinfo{title}{Rethinking atrous convolution for semantic image
  segmentation},
\newblock \bibinfo{journal}{arXiv preprint arXiv:1706.05587}
  (\bibinfo{year}{2017}).
\bibitem[{Chen et~al.(2015)Chen, Papandreou, Kokkinos, Murphy, and
  Yuille}]{chen2014semantic}
\bibinfo{author}{L.-C. Chen}, \bibinfo{author}{G.~Papandreou},
  \bibinfo{author}{I.~Kokkinos}, \bibinfo{author}{K.~Murphy},
  \bibinfo{author}{A.~L. Yuille},
\newblock \bibinfo{title}{Semantic image segmentation with deep convolutional
  nets and fully connected crfs},
\newblock \bibinfo{journal}{ICLR}  (\bibinfo{year}{2015}).
\bibitem[{Wang et~al.(2018)Wang, Girshick, Gupta, and He}]{wang2018non}
\bibinfo{author}{X.~Wang}, \bibinfo{author}{R.~Girshick},
  \bibinfo{author}{A.~Gupta}, \bibinfo{author}{K.~He},
\newblock \bibinfo{title}{Non-local neural networks},
\newblock in: \bibinfo{booktitle}{CVPR}, \bibinfo{year}{2018}, pp.
  \bibinfo{pages}{7794--7803}.
\bibitem[{Cordts et~al.(2016)Cordts, Omran, Ramos, Rehfeld, Enzweiler,
  Benenson, Franke, Roth, and Schiele}]{cordts2016cityscapes}
\bibinfo{author}{M.~Cordts}, \bibinfo{author}{M.~Omran},
  \bibinfo{author}{S.~Ramos}, \bibinfo{author}{T.~Rehfeld},
  \bibinfo{author}{M.~Enzweiler}, \bibinfo{author}{R.~Benenson},
  \bibinfo{author}{U.~Franke}, \bibinfo{author}{S.~Roth},
  \bibinfo{author}{B.~Schiele},
\newblock \bibinfo{title}{The cityscapes dataset for semantic urban scene
  understanding},
\newblock in: \bibinfo{booktitle}{CVPR}, \bibinfo{year}{2016}, pp.
  \bibinfo{pages}{3213--3223}.
\bibitem[{Zhou et~al.(2017)Zhou, Zhao, Puig, Fidler, Barriuso, and
  Torralba}]{zhou2017scene}
\bibinfo{author}{B.~Zhou}, \bibinfo{author}{H.~Zhao},
  \bibinfo{author}{X.~Puig}, \bibinfo{author}{S.~Fidler},
  \bibinfo{author}{A.~Barriuso}, \bibinfo{author}{A.~Torralba},
\newblock \bibinfo{title}{Scene parsing through ade20k dataset},
\newblock in: \bibinfo{booktitle}{CVPR}, \bibinfo{year}{2017}, pp.
  \bibinfo{pages}{633--641}.
\bibitem[{Gong et~al.(2017)Gong, Liang, Zhang, Shen, and Lin}]{gong2017look}
\bibinfo{author}{K.~Gong}, \bibinfo{author}{X.~Liang},
  \bibinfo{author}{D.~Zhang}, \bibinfo{author}{X.~Shen},
  \bibinfo{author}{L.~Lin},
\newblock \bibinfo{title}{Look into person: Self-supervised structure-sensitive
  learning and a new benchmark for human parsing},
\newblock in: \bibinfo{booktitle}{CVPR}, \bibinfo{year}{2017}, pp.
  \bibinfo{pages}{932--940}.
\bibitem[{Lazebnik et~al.(2006)Lazebnik, Schmid, and
  Ponce}]{lazebnik2006beyond}
\bibinfo{author}{S.~Lazebnik}, \bibinfo{author}{C.~Schmid},
  \bibinfo{author}{J.~Ponce},
\newblock \bibinfo{title}{Beyond bags of features: Spatial pyramid matching for
  recognizing natural scene categories},
\newblock in: \bibinfo{booktitle}{CVPR}, volume~\bibinfo{volume}{2},
  \bibinfo{year}{2006}, pp. \bibinfo{pages}{2169--2178}.
\bibitem[{Song et~al.(2018)Song, Wang, Zhao, Shen, and Lam}]{song2018pyramid}
\bibinfo{author}{H.~Song}, \bibinfo{author}{W.~Wang},
  \bibinfo{author}{S.~Zhao}, \bibinfo{author}{J.~Shen}, \bibinfo{author}{K.-M.
  Lam},
\newblock \bibinfo{title}{Pyramid dilated deeper convlstm for video salient
  object detection},
\newblock in: \bibinfo{booktitle}{ECCV}, \bibinfo{year}{2018}, pp.
  \bibinfo{pages}{715--731}.
\bibitem[{Wang et~al.(2019)Wang, Zhao, Shen, Hoi, and Borji}]{wang2019salient}
\bibinfo{author}{W.~Wang}, \bibinfo{author}{S.~Zhao},
  \bibinfo{author}{J.~Shen}, \bibinfo{author}{S.~C. Hoi},
  \bibinfo{author}{A.~Borji},
\newblock \bibinfo{title}{Salient object detection with pyramid attention and
  salient edges},
\newblock in: \bibinfo{booktitle}{CVPR}, \bibinfo{year}{2019}, pp.
  \bibinfo{pages}{1448--1457}.
\bibitem[{Peng et~al.(2017)Peng, Zhang, Yu, Luo, and Sun}]{peng2017large}
\bibinfo{author}{C.~Peng}, \bibinfo{author}{X.~Zhang}, \bibinfo{author}{G.~Yu},
  \bibinfo{author}{G.~Luo}, \bibinfo{author}{J.~Sun},
\newblock \bibinfo{title}{Large kernel matters--improve semantic segmentation
  by global convolutional network},
\newblock in: \bibinfo{booktitle}{CVPR}, \bibinfo{year}{2017}, pp.
  \bibinfo{pages}{4353--4361}.
\bibitem[{Lin et~al.(2017)Lin, Milan, Shen, and Reid}]{lin2017refinenet}
\bibinfo{author}{G.~Lin}, \bibinfo{author}{A.~Milan},
  \bibinfo{author}{C.~Shen}, \bibinfo{author}{I.~Reid},
\newblock \bibinfo{title}{Refinenet: Multi-path refinement networks for
  high-resolution semantic segmentation},
\newblock in: \bibinfo{booktitle}{CVPR}, \bibinfo{year}{2017}, pp.
  \bibinfo{pages}{1925--1934}.
\bibitem[{Ding et~al.(2018)Ding, Jiang, Shuai, Qun~Liu, and
  Wang}]{ding2018context}
\bibinfo{author}{H.~Ding}, \bibinfo{author}{X.~Jiang},
  \bibinfo{author}{B.~Shuai}, \bibinfo{author}{A.~Qun~Liu},
  \bibinfo{author}{G.~Wang},
\newblock \bibinfo{title}{Context contrasted feature and gated multi-scale
  aggregation for scene segmentation},
\newblock in: \bibinfo{booktitle}{CVPR}, \bibinfo{year}{2018}, pp.
  \bibinfo{pages}{2393--2402}.
\bibitem[{Shuai et~al.(2018)Shuai, Zuo, Wang, and Wang}]{shuai2018scene}
\bibinfo{author}{B.~Shuai}, \bibinfo{author}{Z.~Zuo},
  \bibinfo{author}{B.~Wang}, \bibinfo{author}{G.~Wang},
\newblock \bibinfo{title}{Scene segmentation with dag-recurrent neural
  networks},
\newblock \bibinfo{journal}{IEEE TPAMI} \bibinfo{volume}{40}
  (\bibinfo{year}{2018}) \bibinfo{pages}{1480--1493}.
\bibitem[{Zhang et~al.(2018)Zhang, Dana, Shi, Zhang, Wang, Tyagi, and
  Agrawal}]{zhang2018context}
\bibinfo{author}{H.~Zhang}, \bibinfo{author}{K.~Dana},
  \bibinfo{author}{J.~Shi}, \bibinfo{author}{Z.~Zhang},
  \bibinfo{author}{X.~Wang}, \bibinfo{author}{A.~Tyagi},
  \bibinfo{author}{A.~Agrawal},
\newblock \bibinfo{title}{Context encoding for semantic segmentation},
\newblock in: \bibinfo{booktitle}{CVPR}, \bibinfo{year}{2018}, pp.
  \bibinfo{pages}{7151--7160}.
\bibitem[{Liu et~al.(2019)Liu, Chen, Schroff, Adam, Hua, Yuille, and
  Fei-Fei}]{liu2019auto}
\bibinfo{author}{C.~Liu}, \bibinfo{author}{L.-C. Chen},
  \bibinfo{author}{F.~Schroff}, \bibinfo{author}{H.~Adam},
  \bibinfo{author}{W.~Hua}, \bibinfo{author}{A.~L. Yuille},
  \bibinfo{author}{L.~Fei-Fei},
\newblock \bibinfo{title}{Auto-deeplab: Hierarchical neural architecture search
  for semantic image segmentation},
\newblock in: \bibinfo{booktitle}{CVPR}, \bibinfo{year}{2019}, pp.
  \bibinfo{pages}{82--92}.
\bibitem[{Buades et~al.(2005)Buades, Coll, and Morel}]{buades2005non}
\bibinfo{author}{A.~Buades}, \bibinfo{author}{B.~Coll}, \bibinfo{author}{J.-M.
  Morel},
\newblock \bibinfo{title}{A non-local algorithm for image denoising},
\newblock in: \bibinfo{booktitle}{CVPR}, volume~\bibinfo{volume}{2},
  \bibinfo{year}{2005}, pp. \bibinfo{pages}{60--65}.
\bibitem[{Dabov et~al.(2007)Dabov, Foi, Katkovnik, and Egiazarian}]{BM3DTIP07}
\bibinfo{author}{K.~Dabov}, \bibinfo{author}{A.~Foi},
  \bibinfo{author}{V.~Katkovnik}, \bibinfo{author}{K.~Egiazarian},
\newblock \bibinfo{title}{Image denoising by sparse 3-d transform-domain
  collaborative filtering},
\newblock \bibinfo{journal}{IEEE TIP} \bibinfo{volume}{16}
  (\bibinfo{year}{2007}) \bibinfo{pages}{2080--2095}.
\bibitem[{Efros and Leung(1999)}]{efros1999texture}
\bibinfo{author}{A.~A. Efros}, \bibinfo{author}{T.~K. Leung},
\newblock \bibinfo{title}{Texture synthesis by non-parametric sampling},
\newblock in: \bibinfo{booktitle}{ICCV}, volume~\bibinfo{volume}{2},
  \bibinfo{year}{1999}, pp. \bibinfo{pages}{1033--1038}.
\bibitem[{Fu et~al.(2019)Fu, Liu, Tian, Li, Bao, Fang, and Lu}]{fu2019dual}
\bibinfo{author}{J.~Fu}, \bibinfo{author}{J.~Liu}, \bibinfo{author}{H.~Tian},
  \bibinfo{author}{Y.~Li}, \bibinfo{author}{Y.~Bao}, \bibinfo{author}{Z.~Fang},
  \bibinfo{author}{H.~Lu},
\newblock \bibinfo{title}{Dual attention network for scene segmentation},
\newblock in: \bibinfo{booktitle}{CVPR}, \bibinfo{year}{2019}, pp.
  \bibinfo{pages}{3146--3154}.
\bibitem[{Yuan and Wang(2018)}]{yuan2018ocnet}
\bibinfo{author}{Y.~Yuan}, \bibinfo{author}{J.~Wang},
\newblock \bibinfo{title}{Ocnet: Object context network for scene parsing},
\newblock \bibinfo{journal}{arXiv preprint arXiv:1809.00916}
  (\bibinfo{year}{2018}).
\bibitem[{Chorowski et~al.(2015)Chorowski, Bahdanau, Serdyuk, Cho, and
  Bengio}]{chorowski2015attention}
\bibinfo{author}{J.~K. Chorowski}, \bibinfo{author}{D.~Bahdanau},
  \bibinfo{author}{D.~Serdyuk}, \bibinfo{author}{K.~Cho},
  \bibinfo{author}{Y.~Bengio},
\newblock \bibinfo{title}{Attention-based models for speech recognition},
\newblock in: \bibinfo{booktitle}{NeurIPS}, \bibinfo{year}{2015}, pp.
  \bibinfo{pages}{577--585}.
\bibitem[{Wu et~al.(2020)Wu, Chen, Jing, Hu, Ge, and Ji}]{wu2020dynamic}
\bibinfo{author}{F.~Wu}, \bibinfo{author}{F.~Chen}, \bibinfo{author}{X.-Y.
  Jing}, \bibinfo{author}{C.-H. Hu}, \bibinfo{author}{Q.~Ge},
  \bibinfo{author}{Y.~Ji},
\newblock \bibinfo{title}{Dynamic attention network for semantic segmentation},
\newblock \bibinfo{journal}{Neurocomputing} \bibinfo{volume}{384}
  (\bibinfo{year}{2020}) \bibinfo{pages}{182--191}.
\bibitem[{Xu et~al.(2015)Xu, Ba, Kiros, Cho, Courville, Salakhudinov, Zemel,
  and Bengio}]{xu2015show}
\bibinfo{author}{K.~Xu}, \bibinfo{author}{J.~Ba}, \bibinfo{author}{R.~Kiros},
  \bibinfo{author}{K.~Cho}, \bibinfo{author}{A.~Courville},
  \bibinfo{author}{R.~Salakhudinov}, \bibinfo{author}{R.~Zemel},
  \bibinfo{author}{Y.~Bengio},
\newblock \bibinfo{title}{Show, attend and tell: Neural image caption
  generation with visual attention},
\newblock in: \bibinfo{booktitle}{ICML}, \bibinfo{year}{2015}, pp.
  \bibinfo{pages}{2048--2057}.
\bibitem[{Xu and Saenko(2016)}]{xu2016ask}
\bibinfo{author}{H.~Xu}, \bibinfo{author}{K.~Saenko},
\newblock \bibinfo{title}{Ask, attend and answer: Exploring question-guided
  spatial attention for visual question answering},
\newblock in: \bibinfo{booktitle}{ECCV}, \bibinfo{organization}{Springer},
  \bibinfo{year}{2016}, pp. \bibinfo{pages}{451--466}.
\bibitem[{Wang et~al.(2017)Wang, Jiang, Qian, Yang, Li, Zhang, Wang, and
  Tang}]{wang2017residual}
\bibinfo{author}{F.~Wang}, \bibinfo{author}{M.~Jiang},
  \bibinfo{author}{C.~Qian}, \bibinfo{author}{S.~Yang},
  \bibinfo{author}{C.~Li}, \bibinfo{author}{H.~Zhang},
  \bibinfo{author}{X.~Wang}, \bibinfo{author}{X.~Tang},
\newblock \bibinfo{title}{Residual attention network for image classification},
\newblock in: \bibinfo{booktitle}{CVPR}, \bibinfo{year}{2017}, pp.
  \bibinfo{pages}{3156--3164}.
\bibitem[{Hu et~al.(2018)Hu, Shen, and Sun}]{hu2018squeeze}
\bibinfo{author}{J.~Hu}, \bibinfo{author}{L.~Shen}, \bibinfo{author}{G.~Sun},
\newblock \bibinfo{title}{Squeeze-and-excitation networks},
\newblock in: \bibinfo{booktitle}{CVPR}, \bibinfo{year}{2018}, pp.
  \bibinfo{pages}{7132--7141}.
\bibitem[{Yan et~al.(2019)Yan, Wang, Li, Zhang, and Yang}]{yan2019traffic}
\bibinfo{author}{M.~Yan}, \bibinfo{author}{J.~Wang}, \bibinfo{author}{J.~Li},
  \bibinfo{author}{K.~Zhang}, \bibinfo{author}{Z.~Yang},
\newblock \bibinfo{title}{Traffic scene semantic segmentation using
  self-attention mechanism and bi-directional gru to correlate context},
\newblock \bibinfo{journal}{Neurocomputing}  (\bibinfo{year}{2019}).
\bibitem[{Chen et~al.(2016)Chen, Yang, Wang, Xu, and
  Yuille}]{chen2016attention}
\bibinfo{author}{L.-C. Chen}, \bibinfo{author}{Y.~Yang},
  \bibinfo{author}{J.~Wang}, \bibinfo{author}{W.~Xu}, \bibinfo{author}{A.~L.
  Yuille},
\newblock \bibinfo{title}{Attention to scale: Scale-aware semantic image
  segmentation},
\newblock in: \bibinfo{booktitle}{CVPR}, \bibinfo{year}{2016}, pp.
  \bibinfo{pages}{3640--3649}.
\bibitem[{Kong and Fowlkes(2018)}]{kong2018recurrent}
\bibinfo{author}{S.~Kong}, \bibinfo{author}{C.~C. Fowlkes},
\newblock \bibinfo{title}{Recurrent scene parsing with perspective
  understanding in the loop},
\newblock in: \bibinfo{booktitle}{CVPR}, \bibinfo{year}{2018}, pp.
  \bibinfo{pages}{956--965}.
\bibitem[{Kong and Fowlkes(2019)}]{kong2019pixel}
\bibinfo{author}{S.~Kong}, \bibinfo{author}{C.~Fowlkes},
\newblock \bibinfo{title}{Pixel-wise attentional gating for scene parsing},
\newblock in: \bibinfo{booktitle}{WACV}, \bibinfo{year}{2019}, pp.
  \bibinfo{pages}{1024--1033}.
\bibitem[{Pang et~al.(2019)Pang, Li, Shen, and Shao}]{pang2019towards}
\bibinfo{author}{Y.~Pang}, \bibinfo{author}{Y.~Li}, \bibinfo{author}{J.~Shen},
  \bibinfo{author}{L.~Shao},
\newblock \bibinfo{title}{Towards bridging semantic gap to improve semantic
  segmentation},
\newblock in: \bibinfo{booktitle}{ICCV}, \bibinfo{year}{2019}, pp.
  \bibinfo{pages}{4230--4239}.
\bibitem[{Li et~al.(2018)Li, Xiong, An, and Wang}]{li2018pyramid}
\bibinfo{author}{H.~Li}, \bibinfo{author}{P.~Xiong}, \bibinfo{author}{J.~An},
  \bibinfo{author}{L.~Wang},
\newblock \bibinfo{title}{Pyramid attention network for semantic segmentation},
\newblock \bibinfo{journal}{BMVC}  (\bibinfo{year}{2018}).
\bibitem[{Zhu et~al.(2019)Zhu, Xu, Bai, Huang, and Bai}]{zhu2019asymmetric}
\bibinfo{author}{Z.~Zhu}, \bibinfo{author}{M.~Xu}, \bibinfo{author}{S.~Bai},
  \bibinfo{author}{T.~Huang}, \bibinfo{author}{X.~Bai},
\newblock \bibinfo{title}{Asymmetric non-local neural networks for semantic
  segmentation},
\newblock in: \bibinfo{booktitle}{ICCV}, \bibinfo{year}{2019}, pp.
  \bibinfo{pages}{593--602}.
\bibitem[{Huang et~al.(2019)Huang, Wang, Huang, Huang, Wei, and
  Liu}]{huang2019ccnet}
\bibinfo{author}{Z.~Huang}, \bibinfo{author}{X.~Wang},
  \bibinfo{author}{L.~Huang}, \bibinfo{author}{C.~Huang},
  \bibinfo{author}{Y.~Wei}, \bibinfo{author}{W.~Liu},
\newblock \bibinfo{title}{Ccnet: Criss-cross attention for semantic
  segmentation},
\newblock in: \bibinfo{booktitle}{ICCV}, \bibinfo{year}{2019}, pp.
  \bibinfo{pages}{603--612}.
\bibitem[{Zhou et~al.(2019)Zhou, Sun, Zha, and Zeng}]{zhou2019context}
\bibinfo{author}{Y.~Zhou}, \bibinfo{author}{X.~Sun}, \bibinfo{author}{Z.-J.
  Zha}, \bibinfo{author}{W.~Zeng},
\newblock \bibinfo{title}{Context-reinforced semantic segmentation},
\newblock in: \bibinfo{booktitle}{CVPR}, \bibinfo{year}{2019}, pp.
  \bibinfo{pages}{4046--4055}.
\bibitem[{Cao et~al.(2019)Cao, Pang, and Li}]{cao2019triply}
\bibinfo{author}{J.~Cao}, \bibinfo{author}{Y.~Pang}, \bibinfo{author}{X.~Li},
\newblock \bibinfo{title}{Triply supervised decoder networks for joint
  detection and segmentation},
\newblock in: \bibinfo{booktitle}{CVPR}, \bibinfo{year}{2019}, pp.
  \bibinfo{pages}{7392--7401}.
\bibitem[{Dai et~al.(2017)Dai, Qi, Xiong, Li, Zhang, Hu, and
  Wei}]{dai2017deformable}
\bibinfo{author}{J.~Dai}, \bibinfo{author}{H.~Qi}, \bibinfo{author}{Y.~Xiong},
  \bibinfo{author}{Y.~Li}, \bibinfo{author}{G.~Zhang}, \bibinfo{author}{H.~Hu},
  \bibinfo{author}{Y.~Wei},
\newblock \bibinfo{title}{Deformable convolutional networks},
\newblock in: \bibinfo{booktitle}{ICCV}, \bibinfo{year}{2017}, pp.
  \bibinfo{pages}{764--773}.
\bibitem[{Jia et~al.(2016)Jia, De~Brabandere, Tuytelaars, and
  Gool}]{jia2016dynamic}
\bibinfo{author}{X.~Jia}, \bibinfo{author}{B.~De~Brabandere},
  \bibinfo{author}{T.~Tuytelaars}, \bibinfo{author}{L.~V. Gool},
\newblock \bibinfo{title}{Dynamic filter networks},
\newblock in: \bibinfo{booktitle}{NeurIPS}, \bibinfo{year}{2016}, pp.
  \bibinfo{pages}{667--675}.
\bibitem[{Su et~al.(2019)Su, Jampani, Sun, Gallo, Learned-Miller, and
  Kautz}]{su2019pixel}
\bibinfo{author}{H.~Su}, \bibinfo{author}{V.~Jampani},
  \bibinfo{author}{D.~Sun}, \bibinfo{author}{O.~Gallo},
  \bibinfo{author}{E.~Learned-Miller}, \bibinfo{author}{J.~Kautz},
\newblock \bibinfo{title}{Pixel-adaptive convolutional neural networks},
\newblock in: \bibinfo{booktitle}{CVPR}, \bibinfo{year}{2019}, pp.
  \bibinfo{pages}{11166--11175}.
\bibitem[{Han et~al.(2019)Han, Fan, Zhang, and Jiao}]{han2019new}
\bibinfo{author}{H.-H. Han}, \bibinfo{author}{L.~Fan},
  \bibinfo{author}{S.~Zhang}, \bibinfo{author}{D.~Jiao},
\newblock \bibinfo{title}{A new semantic segmentation model for supplementing
  more spatial information},
\newblock \bibinfo{journal}{IEEE Access}  (\bibinfo{year}{2019}).
\bibitem[{Rota~Bul{\`o} et~al.(2018)Rota~Bul{\`o}, Porzi, and
  Kontschieder}]{rota2018place}
\bibinfo{author}{S.~Rota~Bul{\`o}}, \bibinfo{author}{L.~Porzi},
  \bibinfo{author}{P.~Kontschieder},
\newblock \bibinfo{title}{In-place activated batchnorm for memory-optimized
  training of dnns},
\newblock in: \bibinfo{booktitle}{CVPR}, \bibinfo{year}{2018}, pp.
  \bibinfo{pages}{5639--5647}.
\bibitem[{Wu et~al.(2016)Wu, Shen, and Hengel}]{wu2016high}
\bibinfo{author}{Z.~Wu}, \bibinfo{author}{C.~Shen}, \bibinfo{author}{A.~v.~d.
  Hengel},
\newblock \bibinfo{title}{High-performance semantic segmentation using very
  deep fully convolutional networks},
\newblock \bibinfo{journal}{arXiv preprint arXiv:1604.04339}
  (\bibinfo{year}{2016}).
\bibitem[{Liu et~al.(2018)Liu, Ruan, Huang, Wei, Wei, Zhao, and
  Huang}]{liu2018devil}
\bibinfo{author}{T.~Liu}, \bibinfo{author}{T.~Ruan},
  \bibinfo{author}{Z.~Huang}, \bibinfo{author}{Y.~Wei},
  \bibinfo{author}{S.~Wei}, \bibinfo{author}{Y.~Zhao},
  \bibinfo{author}{T.~Huang},
\newblock \bibinfo{title}{Devil in the details: Towards accurate single and
  multiple human parsing},
\newblock \bibinfo{journal}{arXiv preprint arXiv:1809.05996}
  (\bibinfo{year}{2018}).
\bibitem[{He et~al.(2016)He, Zhang, Ren, and Sun}]{he2016deep}
\bibinfo{author}{K.~He}, \bibinfo{author}{X.~Zhang}, \bibinfo{author}{S.~Ren},
  \bibinfo{author}{J.~Sun},
\newblock \bibinfo{title}{Deep residual learning for image recognition},
\newblock in: \bibinfo{booktitle}{CVPR}, \bibinfo{year}{2016}, pp.
  \bibinfo{pages}{770--778}.
\bibitem[{Pohlen et~al.(2017)Pohlen, Hermans, Mathias, and
  Leibe}]{pohlen2017full}
\bibinfo{author}{T.~Pohlen}, \bibinfo{author}{A.~Hermans},
  \bibinfo{author}{M.~Mathias}, \bibinfo{author}{B.~Leibe},
\newblock \bibinfo{title}{Full-resolution residual networks for semantic
  segmentation in street scenes},
\newblock in: \bibinfo{booktitle}{CVPR}, \bibinfo{year}{2017}, pp.
  \bibinfo{pages}{4151--4160}.
\bibitem[{Wang et~al.(2018)Wang, Chen, Yuan, Liu, Huang, Hou, and
  Cottrell}]{wang2018understanding}
\bibinfo{author}{P.~Wang}, \bibinfo{author}{P.~Chen},
  \bibinfo{author}{Y.~Yuan}, \bibinfo{author}{D.~Liu},
  \bibinfo{author}{Z.~Huang}, \bibinfo{author}{X.~Hou},
  \bibinfo{author}{G.~Cottrell},
\newblock \bibinfo{title}{Understanding convolution for semantic segmentation},
\newblock in: \bibinfo{booktitle}{WACV}, \bibinfo{organization}{IEEE},
  \bibinfo{year}{2018}, pp. \bibinfo{pages}{1451--1460}.
\bibitem[{Wu et~al.(2019)Wu, Shen, and Van Den~Hengel}]{wu2019wider}
\bibinfo{author}{Z.~Wu}, \bibinfo{author}{C.~Shen}, \bibinfo{author}{A.~Van
  Den~Hengel},
\newblock \bibinfo{title}{Wider or deeper: Revisiting the resnet model for
  visual recognition},
\newblock \bibinfo{journal}{Pattern Recognition} \bibinfo{volume}{90}
  (\bibinfo{year}{2019}) \bibinfo{pages}{119--133}.
\bibitem[{Liang et~al.(2018)Liang, Zhou, and Xing}]{liang2018dynamic}
\bibinfo{author}{X.~Liang}, \bibinfo{author}{H.~Zhou},
  \bibinfo{author}{E.~Xing},
\newblock \bibinfo{title}{Dynamic-structured semantic propagation network},
\newblock in: \bibinfo{booktitle}{CVPR}, \bibinfo{year}{2018}, pp.
  \bibinfo{pages}{752--761}.
\bibitem[{Yu et~al.(2018{\natexlab{a}})Yu, Wang, Peng, Gao, Yu, and
  Sang}]{yu2018bisenet}
\bibinfo{author}{C.~Yu}, \bibinfo{author}{J.~Wang}, \bibinfo{author}{C.~Peng},
  \bibinfo{author}{C.~Gao}, \bibinfo{author}{G.~Yu}, \bibinfo{author}{N.~Sang},
\newblock \bibinfo{title}{Bisenet: Bilateral segmentation network for real-time
  semantic segmentation},
\newblock in: \bibinfo{booktitle}{ECCV}, \bibinfo{year}{2018}{\natexlab{a}},
  pp. \bibinfo{pages}{325--341}.
\bibitem[{Yu et~al.(2018{\natexlab{b}})Yu, Wang, Peng, Gao, Yu, and
  Sang}]{yu2018learning}
\bibinfo{author}{C.~Yu}, \bibinfo{author}{J.~Wang}, \bibinfo{author}{C.~Peng},
  \bibinfo{author}{C.~Gao}, \bibinfo{author}{G.~Yu}, \bibinfo{author}{N.~Sang},
\newblock \bibinfo{title}{Learning a discriminative feature network for
  semantic segmentation},
\newblock in: \bibinfo{booktitle}{CVPR}, \bibinfo{year}{2018}{\natexlab{b}},
  pp. \bibinfo{pages}{1857--1866}.
\bibitem[{Zhao et~al.(2018)Zhao, Zhang, Liu, Shi, Change~Loy, Lin, and
  Jia}]{zhao2018psanet}
\bibinfo{author}{H.~Zhao}, \bibinfo{author}{Y.~Zhang},
  \bibinfo{author}{S.~Liu}, \bibinfo{author}{J.~Shi},
  \bibinfo{author}{C.~Change~Loy}, \bibinfo{author}{D.~Lin},
  \bibinfo{author}{J.~Jia},
\newblock \bibinfo{title}{Psanet: Point-wise spatial attention network for
  scene parsing},
\newblock in: \bibinfo{booktitle}{ECCV}, \bibinfo{year}{2018}, pp.
  \bibinfo{pages}{267--283}.
\bibitem[{Zhang et~al.(2019{\natexlab{a}})Zhang, Chen, Li, Hong, Liu, Ma, Han,
  and Ding}]{zhang2019acfnet}
\bibinfo{author}{F.~Zhang}, \bibinfo{author}{Y.~Chen}, \bibinfo{author}{Z.~Li},
  \bibinfo{author}{Z.~Hong}, \bibinfo{author}{J.~Liu}, \bibinfo{author}{F.~Ma},
  \bibinfo{author}{J.~Han}, \bibinfo{author}{E.~Ding},
\newblock \bibinfo{title}{Acfnet: Attentional class feature network for
  semantic segmentation},
\newblock in: \bibinfo{booktitle}{ICCV}, \bibinfo{year}{2019}{\natexlab{a}},
  pp. \bibinfo{pages}{6798--6807}.
\bibitem[{Zhang et~al.(2019{\natexlab{b}})Zhang, Li, Arnab, Yang, Tong, and
  Torr}]{zhang2019dual}
\bibinfo{author}{L.~Zhang}, \bibinfo{author}{X.~Li},
  \bibinfo{author}{A.~Arnab}, \bibinfo{author}{K.~Yang},
  \bibinfo{author}{Y.~Tong}, \bibinfo{author}{P.~H. Torr},
\newblock \bibinfo{title}{Dual graph convolutional network for semantic
  segmentation},
\newblock \bibinfo{journal}{BMVC}  (\bibinfo{year}{2019}{\natexlab{b}}).
\bibitem[{Huang et~al.(2017)Huang, Liu, Van Der~Maaten, and
  Weinberger}]{huang2017densely}
\bibinfo{author}{G.~Huang}, \bibinfo{author}{Z.~Liu}, \bibinfo{author}{L.~Van
  Der~Maaten}, \bibinfo{author}{K.~Q. Weinberger},
\newblock \bibinfo{title}{Densely connected convolutional networks},
\newblock in: \bibinfo{booktitle}{CVPR}, \bibinfo{year}{2017}, pp.
  \bibinfo{pages}{4700--4708}.
\bibitem[{Xiao et~al.(2018)Xiao, Liu, Zhou, Jiang, and Sun}]{xiao2018unified}
\bibinfo{author}{T.~Xiao}, \bibinfo{author}{Y.~Liu}, \bibinfo{author}{B.~Zhou},
  \bibinfo{author}{Y.~Jiang}, \bibinfo{author}{J.~Sun},
\newblock \bibinfo{title}{Unified perceptual parsing for scene understanding},
\newblock in: \bibinfo{booktitle}{ECCV}, \bibinfo{year}{2018}, pp.
  \bibinfo{pages}{418--434}.
\bibitem[{Zhang et~al.(2019)Zhang, Zhang, Wang, and Xie}]{zhang2019co}
\bibinfo{author}{H.~Zhang}, \bibinfo{author}{H.~Zhang},
  \bibinfo{author}{C.~Wang}, \bibinfo{author}{J.~Xie},
\newblock \bibinfo{title}{Co-occurrent features in semantic segmentation},
\newblock in: \bibinfo{booktitle}{CVPR}, \bibinfo{year}{2019}, pp.
  \bibinfo{pages}{548--557}.
\bibitem[{He et~al.(2019)He, Deng, Zhou, Wang, and Qiao}]{he2019adaptive}
\bibinfo{author}{J.~He}, \bibinfo{author}{Z.~Deng}, \bibinfo{author}{L.~Zhou},
  \bibinfo{author}{Y.~Wang}, \bibinfo{author}{Y.~Qiao},
\newblock \bibinfo{title}{Adaptive pyramid context network for semantic
  segmentation},
\newblock in: \bibinfo{booktitle}{CVPR}, \bibinfo{year}{2019}, pp.
  \bibinfo{pages}{7519--7528}.
\bibitem[{Wang et~al.(2019)Wang, Zhang, Qi, Shen, Pang, and
  Shao}]{wang2019learning}
\bibinfo{author}{W.~Wang}, \bibinfo{author}{Z.~Zhang}, \bibinfo{author}{S.~Qi},
  \bibinfo{author}{J.~Shen}, \bibinfo{author}{Y.~Pang},
  \bibinfo{author}{L.~Shao},
\newblock \bibinfo{title}{Learning compositional neural information fusion for
  human parsing},
\newblock in: \bibinfo{booktitle}{ICCV}, \bibinfo{year}{2019}, pp.
  \bibinfo{pages}{5703--5713}.
\bibitem[{Wang et~al.(2020)Wang, Zhu, Dai, Pang, Shen, and
  Shao}]{wang2020hierarchical}
\bibinfo{author}{W.~Wang}, \bibinfo{author}{H.~Zhu}, \bibinfo{author}{J.~Dai},
  \bibinfo{author}{Y.~Pang}, \bibinfo{author}{J.~Shen},
  \bibinfo{author}{L.~Shao},
\newblock \bibinfo{title}{Hierarchical human parsing with typed part-relation
  reasoning},
\newblock in: \bibinfo{booktitle}{CVPR}, \bibinfo{year}{2020}, pp.
  \bibinfo{pages}{8929--8939}.
\bibitem[{Liang et~al.(2018)Liang, Gong, Shen, and Lin}]{liang2018look}
\bibinfo{author}{X.~Liang}, \bibinfo{author}{K.~Gong},
  \bibinfo{author}{X.~Shen}, \bibinfo{author}{L.~Lin},
\newblock \bibinfo{title}{Look into person: Joint body parsing \& pose
  estimation network and a new benchmark},
\newblock \bibinfo{journal}{IEEE TPAMI}  (\bibinfo{year}{2018}).
\bibitem[{Zhao et~al.(2017)Zhao, Li, Nie, Zhao, Chen, Wang, Feng, and
  Yan}]{zhao2017self}
\bibinfo{author}{J.~Zhao}, \bibinfo{author}{J.~Li}, \bibinfo{author}{X.~Nie},
  \bibinfo{author}{F.~Zhao}, \bibinfo{author}{Y.~Chen},
  \bibinfo{author}{Z.~Wang}, \bibinfo{author}{J.~Feng},
  \bibinfo{author}{S.~Yan},
\newblock \bibinfo{title}{Self-supervised neural aggregation networks for human
  parsing},
\newblock in: \bibinfo{booktitle}{CVPR Workshops}, \bibinfo{year}{2017}, pp.
  \bibinfo{pages}{7--15}.
\bibitem[{Luo et~al.(2018)Luo, Zheng, Zheng, Guan, Yu, and Yang}]{luo2018macro}
\bibinfo{author}{Y.~Luo}, \bibinfo{author}{Z.~Zheng},
  \bibinfo{author}{L.~Zheng}, \bibinfo{author}{T.~Guan},
  \bibinfo{author}{J.~Yu}, \bibinfo{author}{Y.~Yang},
\newblock \bibinfo{title}{Macro-micro adversarial network for human parsing},
\newblock in: \bibinfo{booktitle}{ECCV}, \bibinfo{year}{2018}, pp.
  \bibinfo{pages}{418--434}.
\bibitem[{Nie et~al.(2018)Nie, Feng, and Yan}]{nie2018mutual}
\bibinfo{author}{X.~Nie}, \bibinfo{author}{J.~Feng}, \bibinfo{author}{S.~Yan},
\newblock \bibinfo{title}{Mutual learning to adapt for joint human parsing and
  pose estimation},
\newblock in: \bibinfo{booktitle}{ECCV}, \bibinfo{year}{2018}, pp.
  \bibinfo{pages}{502--517}.

\end{thebibliography}

\end{document}